%% file: main.tex
\documentclass[lettersize,journal]{IEEEtran}
\usepackage{amsmath,amsfonts}
\usepackage{algorithmic}
\usepackage{algorithm}
\usepackage{array}
\usepackage[caption=false,font=normalsize,labelfont=sf,textfont=sf]{subfig}
\usepackage{textcomp}
\usepackage{stfloats}
\usepackage{url}
\usepackage{verbatim}
\usepackage{graphicx}
\usepackage{cite}
\usepackage[colorlinks, linkcolor=blue, anchorcolor=blue, citecolor=blue]{hyperref}

\usepackage{booktabs}
\usepackage{tabularx}
\usepackage{float}
\usepackage{multirow}
\usepackage{makecell}
\makeatletter
\let\NAT@parse\undefined
\makeatother
\begin{document}

\title{A Generative Data Framework with Reliable Supervision for Underwater Image Restoration and Enhancement}

\author{
    Yufeng Tian$^*$, Yifan Chen$^*$, Zhe Sun$^{*,\dagger}$, Libang Chen, Mingyu Dou, Jijun Lu, Ye Zheng, and 
    \\ Xuelong Li$^{\dagger}$,~\IEEEmembership{Fellow,~IEEE}


    \thanks{Yufeng Tian, Zhe Sun, Libang Chen, Jijun Lu, Ye Zheng, and Xuelong Li are with the Institute of Artificial Intelligence (TeleAI), China Telecom, Shanghai 200232, China (e-mail: tian\_yufeng@outlook.com; sunzhe@nwpu.edu.cn; chenlb35@mail3.sysu.edu.cn; jijun\_lu@163.com; zhengye@westlake.edu.cn; xuelong\_li@ieee.org).}

    \thanks{Yifan Chen is with the Institute of Artificial Intelligence (TeleAI), China Telecom, Shanghai 200232, China, and also with the College of Future Information Technology, Fudan University, Shanghai 200433, China (e-mail: chenyifan@fudan.edu.cn).}

    \thanks{Mingyu Dou is with the Xi'an Institute of Optics and Precision Mechanics, Chinese Academy of Sciences, Xi'an 710119, China (e-mail: doumingyu24@mails.ucas.ac.cn).}
    
    \thanks{$^*$These authors contributed equally to this work.}
    \thanks{$^{\dagger}$Corresponding authors: Zhe Sun and Xuelong Li.}
}



\maketitle

\begin{abstract}
Learning-based underwater image restoration and enhancement rely heavily on high-quality paired training data.
However, obtaining clear reference images for real underwater scenes remains challenging.
Existing datasets often use the outputs of enhancement algorithms as references, which may introduce uncertainty into supervision targets.
To address this issue, we propose a generative data framework that preserves clear source images as restoration references and synthesizes corresponding underwater inputs through unpaired image-to-image translation.
Using representative underwater appearances to guide the translation, we construct UWNature from natural photographs.
We systematically examine the reliability of the generated supervision through its agreement with real underwater appearances, structural consistency, and depth-related degradation characteristics.
Experiments across 6 restoration architectures and 3 test sets evaluate its effectiveness through objective metrics, human perceptual assessment, and the correspondence between synthetic degradation and learned restoration behavior.
The results demonstrate competitive restoration performance and favorable perceptual quality on real underwater images.
Further analysis reveals that greater spatial plausibility of synthetic degradation does not necessarily improve restoration performance, highlighting the importance of evaluating both the intrinsic properties of synthetic data and their effectiveness for model learning.
These findings support a scalable framework for constructing and evaluating reliable paired supervision for underwater image restoration and enhancement.
The generated datasets are available at: \url{https://github.com/tianyufengcn/SynUIEDatasets}.
\end{abstract}


\input{sections/01_introduction}

\input{sections/02_related_works}

\input{sections/03_framework}

\input{sections/04_intrinsic}

\input{sections/05_downstream}

\input{sections/06_conclusion}

\bibliographystyle{IEEEtran}
\bibliography{ref}

\clearpage
\appendix
\input{appendix/supplement}

\vfill

\end{document}

%% file: sections/01_introduction.tex
\section{Introduction}

\IEEEPARstart{U}{nderwater} images often suffer from color distortion, low contrast, and blurred details due to light absorption and scattering in water \cite{raveendran2021underwater},\cite{saoud2024seeing},\cite{akkaynak2018formation}.
These degradations reduce image visibility and affect subsequent underwater vision tasks \cite{cong2024comprehensive},\cite{mohsan2023recent}.
In recent years, deep learning has significantly advanced underwater image restoration and enhancement \cite{li2020underwater},\cite{peng2023u},\cite{zhao2024wavelet}.
However, restoring image quality across diverse underwater conditions remains challenging.

The performance of deep learning models largely depends on the quality of their training data.
Obtaining underwater images with reliable clear references remains difficult.
Many existing underwater image datasets use the outputs of enhancement algorithms as references, often selected by human observers \cite{li2019underwater},\cite{peng2023u}.
These references may contain color and detail biases introduced by the algorithms or subjective selection~\cite{liu2025pseudoref}.
Synthetic data provide another solution by generating degraded images from clear sources without collecting paired underwater images~\cite{li2018watergan},\cite{li2025muse}.
However, synthetic degradation may differ from real underwater conditions~\cite{10048777} and may introduce changes to the original image structure.

The key to reliable synthetic supervision is not simply generating paired images, but ensuring the quality of the pairs.
The reference should provide a clear and consistent restoration target, while the generated image should reproduce plausible underwater degradation and preserve the original scene structure.
Moreover, the reliability of synthetic supervision should be examined through both the properties of the generated data and the restoration performance of models trained on them.

To address these challenges, we propose a generative data framework that uses clear source images as restoration references and synthesizes corresponding underwater degradation through unpaired image-to-image translation\cite{zhu2017unpaired}.
The original clear images are preserved as references, while the generated images serve as degraded inputs for supervised learning.
Using  representative underwater appearances as translation targets, we instantiate the framework with natural photographs to construct UWNature.
The same framework is also applied to AI-generated reference images to construct UWImgNetSD.
This approach provides reliable reference targets and diverse synthetic degradation for underwater image restoration.

To evaluate the reliability of the proposed framework, we examine the intrinsic properties of the generated supervision using UWNature as the main instance and assess the restoration models trained on the synthetic datasets.
We first compare real and synthetic underwater appearances, measure the structural consistency of the image pairs, and analyze how degradation varies with relative scene depth.
Based on this analysis, we further introduce depth-aware modulation to adjust the spatial characteristics of synthetic degradation.
We then investigate the correspondence between synthetic degradation and learned restoration behavior and evaluate the trained models using multiple objective metrics and a human perceptual study.
These evaluations demonstrate the effectiveness of synthetic supervision while revealing how the characteristics of generated degradation influence restoration performance.

The main contributions of this work are summarized as follows:

\begin{itemize}

\item We propose a generative data framework that preserves clear images as restoration references and synthesizes underwater degradation to construct reliable paired supervision.
Using this framework, we construct UWNature from natural photographs and UWImgNetSD from AI-generated images, demonstrating its applicability to different clear image sources.

\item We systematically evaluate the intrinsic properties of synthetic supervision using UWNature, examining its appearance distribution relative to real underwater images, structural consistency, and spatial degradation characteristics.
We further introduce depth-aware modulation to adjust the degradation behavior identified in the original synthesis.

\item We investigate the restoration behavior learned from synthetic supervision and evaluate its effectiveness across  restoration architectures and 3 test sets using multiple objective metrics and a human perceptual study.
The results demonstrate the effectiveness of the generated supervision and show that improving the spatial plausibility of synthetic degradation does not necessarily improve restoration performance across architectures and metrics.

\end{itemize}

%% file: sections/02_related_works.tex
\section{Related Work}

\subsection{Underwater Image Datasets}

The development of deep learning methods for underwater image Restoration and Enhancement relies on suitable training datasets.
UIEB~\cite{li2019underwater} provides real underwater images with reference images selected through human evaluation of different enhancement results.
LSUI~\cite{peng2023u} expands the scale and diversity of underwater training data, while EUVP~\cite{islam2020fast} provides paired and unpaired images with different levels of visual quality.
These datasets have supported the development of many underwater restoration methods.

However, obtaining clear references for real underwater scenes remains difficult.
References selected from enhancement results reflect the characteristics of the algorithms and the preferences of human observers~\cite{liu2025pseudoref}.
Fu et al.~\cite{fu2022uncertainty} addressed the ambiguity of reference images by modeling underwater enhancement as a probabilistic distribution rather than a deterministic mapping.
These studies highlight the importance of reference quality in supervised restoration.
In our framework, the original clear images serve directly as reference targets, while their underwater degraded counterparts are generated as training inputs.

\subsection{Synthetic Underwater Image Generation}
Synthetic data provide another approach to obtaining paired training images.
Physics-based methods model the effects of light absorption and scattering in water~\cite{akkaynak2018formation},\cite{li2025muse}.
WaterGAN~\cite{li2018watergan} uses natural images and depth information to generate underwater images for training color correction models.
Sea-thru~\cite{akkaynak2019seathru} further demonstrates the importance of scene distance and underwater light propagation in recovering image colors.
Such methods provide a physical basis for understanding underwater degradation, although their application depends on the available scene information and assumptions about water conditions.

Learning-based methods provide greater flexibility in generating underwater appearances~\cite{zhang2024atlantis}.
Generative adversarial networks and image translation methods can learn visual characteristics from real underwater images without requiring paired underwater references~\cite{fabbri2018enhancing},\cite{zhu2017unpaired}.
Domain adaptation methods have also been explored to reduce the differences between synthetic and real underwater images.
Jiang et al.~\cite{JIANG2022108324} proposed a two-step approach that transfers image restoration knowledge from the air domain to underwater scenes.
Wang et al.~\cite{10048777} further investigated domain differences in underwater enhancement.
These studies demonstrate the value of learning from different image domains.
Our work uses image translation to construct paired supervision and examines the appearance and structural consistency of the generated training pairs.

\subsection{Evaluation of Underwater Restoration and Enhancement}
Evaluating underwater restoration requires considering both image quality and the availability of reference images~\cite{zheng2022uif},\cite{jiang2024towards},\cite{jiang2025degradationiqa}.
Full-reference metrics, including PSNR, SSIM~\cite{1284395}, and LPIPS~\cite{8578166}, measure differences from a provided reference.
For real underwater images without clear references, UCIQE~\cite{7300447} and UIQM~\cite{7305804} are widely used to assess visual attributes such as color, contrast, and sharpness.
Recent methods, including PAUQA~\cite{11017762}, MUSIQ~\cite{9710973}, TOPIQ~\cite{10478301}, LIQE~\cite{10204315}, and CLIP-IQA+~\cite{10.1609/aaai.v37i2.25353}, provide additional assessments of perceptual image quality.
Human subjective studies offer another means of evaluating overall visual quality~\cite{guo2022subjectiveiqa},\cite{jiang2022underwater}.

These measures provide useful evidence of restoration performance.
Our work further examines the agreement between real and generated underwater appearances, the structural consistency of training pairs, and the relationship between synthetic degradation and learned restoration behavior.
Together, these analyses and downstream evaluations assess the reliability and effectiveness of the proposed synthetic supervision.

%% file: sections/03_framework.tex
\section{Generative Data Framework}
\label{sec:framework}

\subsection{Overview}
\label{sec:framework_overview}

\begin{figure*}[t]
    \centering
    \includegraphics[width=0.98\textwidth]{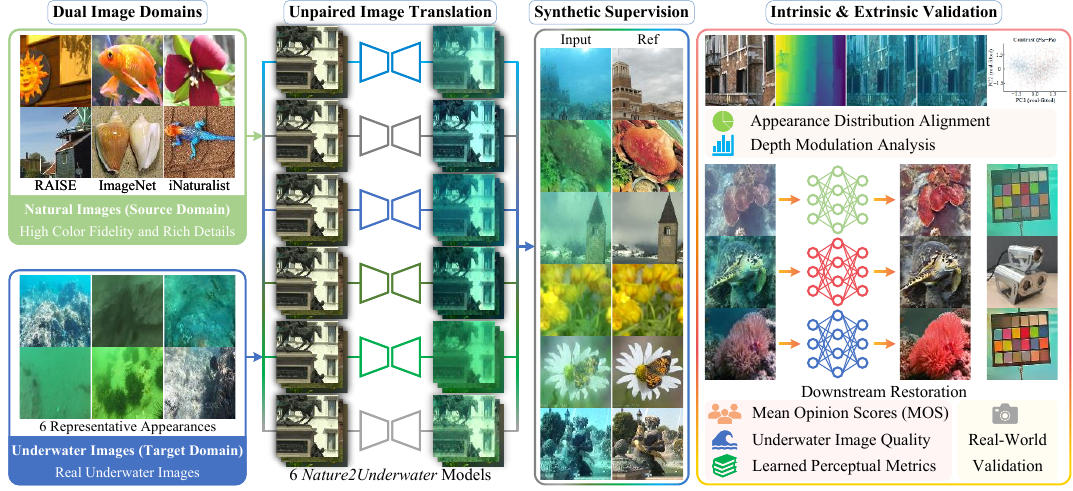}
    \caption{Overview of the proposed generative data framework.
    6 representative underwater appearances guide unpaired image translation to construct synthetic degraded inputs paired with their original clear references.
    The resulting supervision is examined through intrinsic data analysis, downstream restoration performance, human perceptual evaluation, and real-world validation.}
    \label{fig:framework}
\end{figure*}

This section presents the generative data framework used to construct reliable supervision for underwater image restoration and enhancement.
Specifically, image-to-image translation (I2IT)~\cite{zhu2017unpaired} transforms clear source images into images with underwater appearance characteristics while preserving the original images as reference targets.
Fig.~\ref{fig:framework} shows the overall framework.

$\mathcal{Y}$ and $\mathcal{X}$ denote the clear source image domain and the underwater image domain, respectively.
Given a reference image $y\in\mathcal{Y}$ and a target appearance $a$, the translation model $G_a$ generates an underwater image $x=G_a(y)\in\mathcal{X}$.
The generated image $x$ is paired with its original reference $y$, forming the synthetic supervision dataset
\begin{equation}
    \mathcal{D}
    =
    \left\{
    \left(x^{(i)},y^{(i)}\right)
    \right\}_{i=1}^{N}.
    \label{eq:synthetic_dataset}
\end{equation}
Given these pairs, a restoration network $H_{\theta}$ learns the inverse mapping from the underwater input $x$ to the clear target $y$, i.e., $\hat{y}=H_{\theta}(x)$.

\subsection{Image Domain Preparation}
\label{sec:image_domain_preparation}

To instantiate the framework with UWNature, appropriate data are prepared for both the in-air source domain and the underwater target domain to support image translation and supervision construction.

\subsubsection{Source Domain (In-air Images)}

The source domain consists of high-quality natural images captured in air, with rich texture details and diverse scene content.
Images are selected from 3 public datasets: RAISE~\cite{dang2015raise}, ImageNet~\cite{deng2009imagenet}, and iNaturalist-12K~\cite{van2018inaturalist}.
RAISE provides high-resolution natural photographs captured in real-world scenes.
ImageNet provides broad diversity in objects and scene content.
iNaturalist-12K contains real-world photographs of plants and animals and further enriches the natural content of the source domain.
Together, the 3 datasets form the in-air image domain
\begin{equation}
    \mathcal{Y}
    =
    \mathcal{Y}_{\mathrm{RAISE}}
    \cup
    \mathcal{Y}_{\mathrm{ImageNet}}
    \cup
    \mathcal{Y}_{\mathrm{iNaturalist}}.
    \label{eq:source_domain}
\end{equation}

\subsubsection{Target Domain (Underwater Degraded Images)}

The target domain consists of images collected from real underwater environments.
Due to variations in underwater absorption and scattering, these images exhibit diverse visual appearances in color cast, illumination, contrast, and clarity.
Underwater degradations vary continuously, with multiple appearance changes often occurring together.
For I2IT training, 6 visually distinctive and representative underwater appearances are selected as target domains.
The target domain is defined as
\begin{equation}
    \mathcal{X}
    =
    \bigcup_{i=1}^{6}
    \mathcal{X}_{a_i},
    \label{eq:target_domain}
\end{equation}
where $a_i$ denotes the $i$-th target appearance.
These target domains serve as appearance anchors that guide the translation models to learn diverse underwater characteristics rather than mutually exclusive physical degradation categories.

\subsection{Nature2Underwater Synthesis and Dataset Construction}
\label{sec:i2it_dataset_construction}

To address the unpaired source and target domains, CycleGAN-Turbo~\cite{parmar2024one} is adopted for Nature2Underwater translation.
CycleGAN-Turbo is an unpaired I2IT method based on a one-step diffusion model.
The official implementation and training paradigm are followed in this work.

As CycleGAN-Turbo learns translation between a source and a target domain, 6 independent translation models are trained for the 6 underwater appearance targets described above.
Each model is specifically trained to translate in-air natural images into a particular underwater appearance.
Using the 6 trained models, source images from RAISE, ImageNet, and iNaturalist-12K are translated into 6 groups of underwater images.
Images sampled from these groups are combined to form the synthetic dataset.

Let $G_{a_i}$ denote the translation model trained for the $i$-th target appearance.
For a chosen source image set $\mathcal{Y}_{\mathrm{src}}$, let $\mathcal{S}_i\subseteq\mathcal{Y}_{\mathrm{src}}$ denote the images selected for translation to appearance $a_i$.
The synthetic supervision dataset is constructed as
\begin{equation}
    \mathcal{D}(\mathcal{Y}_{\mathrm{src}})
    =
    \bigcup_{i=1}^{6}
    \left\{
    (G_{a_i}(y),y)
    \mid y\in\mathcal{S}_i
    \right\}.
    \label{eq:dataset_construction}
\end{equation}
For the natural images defined in Eq.~\eqref{eq:source_domain}, the original clear images are preserved as references, and the resulting dataset is denoted as \emph{UWNature}.

The same framework can be applied to other clear image sources.
To explore AI-generated content (AIGC) as reference images, images from Stable ImageNet-1K are used as source-domain images.
Stable ImageNet-1K contains images synthesized by Stable Diffusion v1.4 using ImageNet-1K class labels as prompts.
The same 6 trained translation models are used to convert these AI-generated images into underwater images, resulting in a second dataset denoted as \emph{UWImgNetSD}.

%% file: sections/04_intrinsic.tex
\section{Intrinsic Validation of Synthetic Supervision}
\label{sec:intrinsic_validation}

This section evaluates the intrinsic properties of synthetic supervision generated by the proposed framework using UWNature as an instance.
The analysis examines the agreement between real and synthetic underwater appearances, the structural consistency of image pairs, and the variation of synthetic degradation with scene depth.
The analysis uses 1,523 paired samples from UWNature and a balanced set of 2,850 real underwater images.

\subsection{Real and Synthetic Underwater Appearance}
\label{sec:distribution_alignment}

Real underwater images exhibit diverse appearances because light absorption and scattering vary across scenes and imaging conditions~\cite{akkaynak2018formation}.
To characterize these variations, each image is represented by the appearance descriptor
\begin{equation}
    \mathbf{f}(x)
    =
    [a^*,b^*,L^*,C,\log_{10}(T+\epsilon)]^{\top},
    \label{eq:appearance_descriptor}
\end{equation}
where $a^*$, $b^*$, and $L^*$ are median CIELab components, $C$ is the difference between the 95th and 5th luminance percentiles, $T$ is the Tenengrad detail measure, and $\epsilon=10^{-12}$ ensures numerical stability.
The real underwater images form continuous distributions across color, luminance, contrast, and detail.
These attributes also vary together: after controlling for the source dataset, low contrast and low detail have a Spearman correlation of 0.5887.
This relationship illustrates the coexistence of multiple appearance changes in real underwater observations.

\begin{figure}[t]
    \centering
    \includegraphics[width=0.48\textwidth]{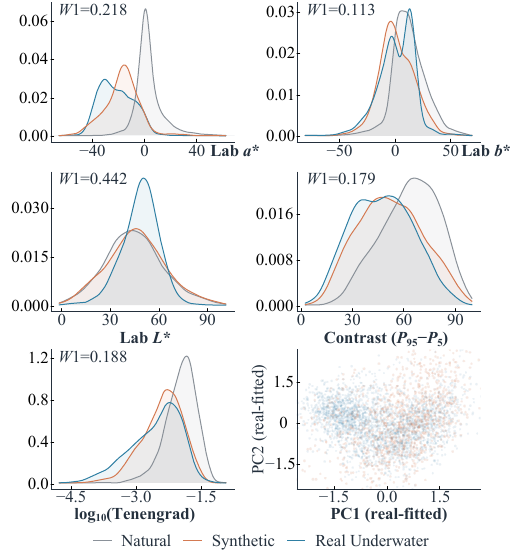}
    \caption{Appearance distributions of natural references, synthetic underwater images, and real underwater images across 5 descriptors, with Wasserstein-1 ($W_1$) distances between real and synthetic distributions indicated.
    The bottom-right panel shows the joint distribution of real and synthetic images in a PCA space fitted to real underwater data.}
    \label{fig:distribution_alignment}
\end{figure}

The changes introduced by synthesis are measured by comparing each generated underwater image $x$ with its in-air reference $y$.
For each pair, the appearance change is defined as $\Delta\mathbf{f}=\mathbf{f}(x)-\mathbf{f}(y)$.
The generated images exhibit consistent shifts along the CIELab color axes, with median changes of $-18.80$ in $a^*$ and $-11.86$ in $b^*$.
Contrast also decreases, while a negative change in the logarithmic Tenengrad measure indicates a reduction in image detail.
The changes in luminance are less directional than the changes in color and detail.

The distributions of real and synthetic underwater images are compared using Wasserstein-1 distance (W1) along each appearance attribute and Sliced Wasserstein Distance (SWD) across the joint descriptor space.
Neighborhood precision and coverage are also measured to evaluate how the synthesized images occupy the real appearance space.

Fig.~\ref{fig:distribution_alignment} compares the appearance distributions of natural references, synthetic underwater images, and real underwater images, together with their joint-space projection.
For the joint-space visualization, PCA is fitted on the balanced real subset and used to project the full collection of 7,760 real underwater images and the synthetic samples.
The generated images follow the main color, contrast, and detail characteristics observed in real underwater data and occupy a substantial part of the real appearance space.
This correspondence is reflected by a precision of 0.8070, coverage of 0.6056, and mean SWD of 0.2044.
The luminance axis shows the largest W1 distance, while the other appearance dimensions exhibit closer agreement.
Together, the paired changes and distribution comparison show that the I2IT process reproduces key appearance variations associated with underwater degradation.

\subsection{Structural Consistency of Synthetic Supervision}
\label{sec:structural_consistency}

A synthetic supervision pair should retain the scene content and spatial arrangement of its reference image so that the restoration model learns the intended appearance correction.
The correspondence between each generated input and its reference is measured using grayscale correlation, edge correlation, and edge SSIM.

\begin{table}[h]
    \centering
    \caption{Structure and depth statistics of UWNature.}
    \label{tab:structure_depth}
    \footnotesize
    \setlength{\tabcolsep}{3pt}
    \renewcommand{\arraystretch}{1.15}
    \begin{tabularx}{0.95\columnwidth}{
        @{}
        >{\raggedright\arraybackslash}X
        c
        >{\raggedright\arraybackslash}X
        c
        @{}
    }
        \toprule
        \multicolumn{2}{c}{\textbf{Pair structure}} &
        \multicolumn{2}{c}{\textbf{Depth association}} \\
        \cmidrule(lr){1-2}\cmidrule(lr){3-4}
        Metric & Median & Attribute & Median $\rho$ \\
        \midrule
        Grayscale correlation & 0.9426 &
        Chroma shift & $-0.1551$ \\
        Edge correlation & 0.8602 &
        $|\Delta L|$ & $-0.0752$ \\
        Edge SSIM & 0.7133 &
        Log contrast ratio & $0.0675$ \\
        & &
        Log detail ratio & $0.0698$ \\
        \bottomrule
    \end{tabularx}
\end{table}

As reported in Table~\ref{tab:structure_depth}, the generated images retain strong correlations with the reference structures.
The combined intensity and edge measurements indicate that synthesis mainly changes image appearance while preserving the underlying spatial structure.

\subsection{Spatial Variation of Degradation with Depth}
\label{sec:depth_analysis}

Light attenuation and scattering depend on the distance traveled through water, so the appearance of an underwater scene can change with depth~\cite{akkaynak2018formation},\cite{akkaynak2019seathru}.
The spatial behavior of the synthetic degradation is examined by relating local image changes to estimated relative scene depth.

Depth Anything V2 Large~\cite{depth_anything_v2} estimates depth from the in-air reference images.
MoGe-2~\cite{NEURIPS2025_336572db} provides an independent estimate for comparison.
The depth estimates are converted into ordinal ranks for each image, with 0 denoting near regions and 1 denoting far regions.
The 2 estimators show strong agreement in relative depth ordering, with a median Spearman correlation of 0.9589 across 1,454 valid matched samples.

For each pair, local changes in chroma, luminance, contrast, and detail are measured against the corresponding depth ranks.
Chroma shift is measured as the Euclidean change in CIELab $a^*b^*$ components, while luminance change is measured as the absolute difference in $L^*$.
The log contrast and detail ratios are computed from local luminance standard deviation and Sobel gradient energy, respectively, using $21\times21$ neighborhoods.
The spatial association between relative depth and each degradation attribute is measured using block-wise Spearman correlation.
Both the depth map and the attribute map are divided into a $16\times16$ grid, and their block medians are used to calculate
\begin{equation}
    \rho_q
    =
    \operatorname{Spearman}
    \left(\{d_k\},\{q_k\}\right),
    \label{eq:depth_correlation}
\end{equation}
where $d_k$ and $q_k$ denote the median depth and degradation attribute in the $k$-th valid block.
The median correlation across images describes the dominant spatial trend of each attribute.

\begin{figure}[t]
    \centering
    \includegraphics[width=0.48\textwidth]{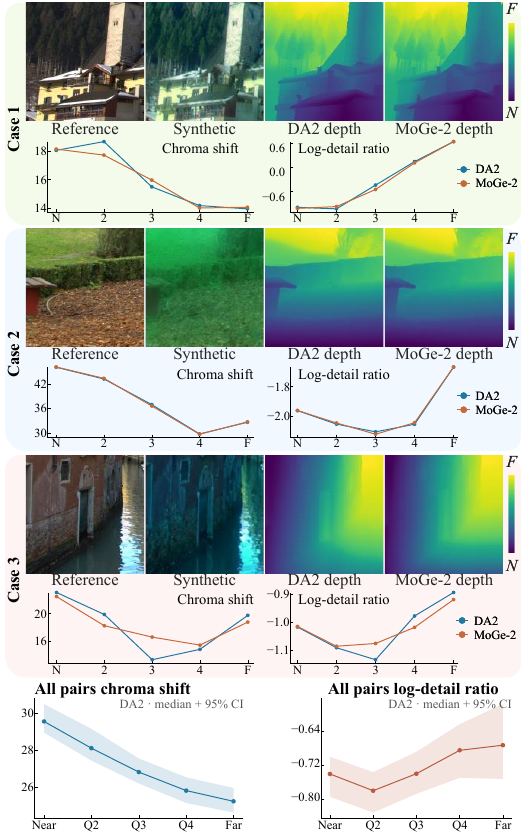}
    \caption{Spatial variation of synthetic degradation across relative scene depth.
    3 representative cases show reference and synthetic images, depth estimates from DA2 and MoGe-2, and their corresponding depth-binned chroma shifts and log-detail ratios.
    The bottom panels summarize the trends across all pairs using DA2, with median values and 95\% confidence intervals.}
    \label{fig:depth_analysis}
\end{figure}

Table~\ref{tab:structure_depth} shows that the original synthesis tends to produce smaller chromatic changes in distant regions.
The positive associations for the log contrast and detail ratios also indicate weaker relative attenuation toward greater depth.
These trends do not follow the expected increase in the effects of light propagation with scene depth.
The same general behavior is obtained with MoGe-2 and remains consistent when the analysis is restricted to images with reliable depth estimates.
The analysis therefore identifies a specific aspect of the original synthesis that can be examined through additional spatial modulation.
Fig.~\ref{fig:depth_analysis} illustrates the estimated depth and the changes in degradation attributes across near and far regions.
The depth-dependent curves are obtained by dividing relative depth into five intervals and computing the median degradation attributes within each interval.

\subsection{Depth Modulation and Analysis}
\label{sec:depth_modulation}

The spatial analysis above shows that the original synthetic degradation does not consistently follow the expected trend with depth.
To examine whether this behavior can be adjusted without retraining the translation models, 3 modulation operators are applied to the synthetic images.
The original input is denoted by $x_O=x$, and its relative depth map $D$ is estimated from the reference $y$.
Only the generated input is modified; the reference remains unchanged.
The relative depth map is smoothed to obtain a shared modulation weight
\begin{equation}
    w(p)
    =
    \operatorname{clip}
    \left(
    \frac{\mathcal{G}_{3}(DM)(p)}
    {\mathcal{G}_{3}(M)(p)},
    0,1
    \right),
    \label{eq:depth_weight}
\end{equation}
where $p$ denotes a spatial position, $M$ is the valid-depth mask, and $\mathcal{G}_{3}$ denotes Gaussian smoothing with a standard deviation of 3 pixels.
The weight $w$ increases from near to far regions and is shared by all modulation operators.

\subsubsection{Modulation Operators}

\noindent (1) Residual Redistribution (R):
The residual redistribution operator varies the strength of the existing degradation residual according to relative depth.
Let $\Delta(p)=x_O(p)-y(p)$ denote the degradation residual introduced by I2IT.
The depth-dependent residual weight is defined as
\begin{equation}
    \alpha(p)
    =
    \operatorname{clip}
    \left[
    1+\lambda(w(p)-\bar{w}),
    0.90,1.10
    \right],
    \label{eq:residual_weight}
\end{equation}
where $\bar{w}$ is the mean depth weight and $\lambda=0.20$ controls the redistribution strength.
The modulated image is then obtained by
\begin{equation}
    x_R(p)
    =
    \operatorname{clip}
    \left[
    y(p)+\alpha(p)\Delta(p),
    0,1
    \right].
    \label{eq:residual_modulation}
\end{equation}
This operation redistributes the existing appearance changes, giving greater weight to regions with larger estimated depth.

\noindent (2) Depth Veil (V):
The veil operator adds a color veil whose strength varies with relative depth.
The ambient color $A_O$ is estimated as the channel-wise median RGB color of the original synthetic image $x_O$ over pixels above the 90th percentile of valid depth weights.
For an input image $z$, the veil operation is expressed as
\begin{equation}
    \begin{aligned}
    \mathcal{V}(z;w,A_O)(p)
    =\operatorname{clip}\big[
    &(1-\beta(p))z(p) 
    +\beta(p)A_O,\;0,1
    \big],
    \end{aligned}
    \label{eq:depth_veil}
\end{equation}
where $\beta(p)=0.10w(p)$ controls the depth-dependent mixing strength.
The mixing strength increases with relative depth, producing a stronger scattering appearance in distant regions.
Applying this operator to the original synthesis gives $x_V=\mathcal{V}(x_O;w,A_O)$.

\noindent (3) Detail Attenuation (B):
The detail operator reduces high-frequency content as relative depth increases.
It blends the input image with its Gaussian-smoothed version using a depth-dependent weight $\beta_B(p)=0.12w(p)$ and a Gaussian standard deviation of 1 pixel.
This provides an additional means of varying detail loss between near and far regions.

These operators are evaluated individually and in combination, producing 8 variants: O, R, V, B, RV, RB, VB, and RVB.
For the combined RV variant, residual redistribution is followed by depth veiling:
\begin{equation}
    x_{RV}
    =
    \mathcal{V}(x_R;w,A_O).
    \label{eq:combined_rv}
\end{equation}
All variants use the same source pairs and depth weights, with operators applied in the order R, V, and B when combined.
Each operator clips its output to $[0,1]$, and the reference images remain unchanged.

\begin{table}[h]
    \centering
    \caption{Intrinsic comparison of 8 modulation variants.}
    \label{tab:modulation_ablation}
    \footnotesize
    \setlength{\tabcolsep}{3pt}
    \renewcommand{\arraystretch}{1.15}
    \begin{tabular*}{0.95\columnwidth}{
        @{}l@{\extracolsep{\fill}}rcccc@{}
    }
        \toprule
        Variant &
        \shortstack{Chroma $\rho$} &
        \shortstack{SWD $\downarrow$} &
        \shortstack{Precision $\uparrow$} &
        \shortstack{Coverage $\uparrow$} &
        \shortstack{Edge SSIM $\uparrow$} \\
        \midrule
        O   & $-0.1551$ & 0.2044 & 0.8070 & 0.6056 & 0.7382 \\
        R   & $0.0180$  & 0.2074 & 0.7932 & 0.6028 & 0.7423 \\
        V   & $-0.1555$ & 0.1857 & 0.8089 & 0.6042 & 0.7271 \\
        B   & $-0.1545$ & 0.2004 & 0.7997 & 0.6000 & 0.7326 \\
        RV  & $0.0115$  & 0.1885 & 0.7997 & 0.6088 & 0.7331 \\
        RB  & $0.0180$  & 0.2034 & 0.7905 & 0.6025 & 0.7375 \\
        VB  & $-0.1555$ & 0.1825 & 0.8070 & 0.6074 & 0.7243 \\
        RVB & $0.0112$  & 0.1852 & 0.8011 & 0.6095 & 0.7287 \\
        \bottomrule
    \end{tabular*}
\end{table}

\subsubsection{Analysis of Modulation Variants}

The variants are compared using the same depth associations, real and synthetic appearance statistics, and structural measurements.
Table~\ref{tab:modulation_ablation} summarizes the results.
For this comparison, edge SSIM is computed from raw Sobel edge responses, while the pair-structure analysis in Table~\ref{tab:structure_depth} uses percentile-normalized edge maps.
R shifts the negative chroma association with depth toward zero but provides little improvement in global distribution alignment.
V improves the spatial trends of contrast and detail and reduces the distribution distance, while its chroma trend remains similar to the original synthesis.
B primarily increases the attenuation of image details.

\begin{figure}[t]
    \centering
    \includegraphics[width=0.48\textwidth]{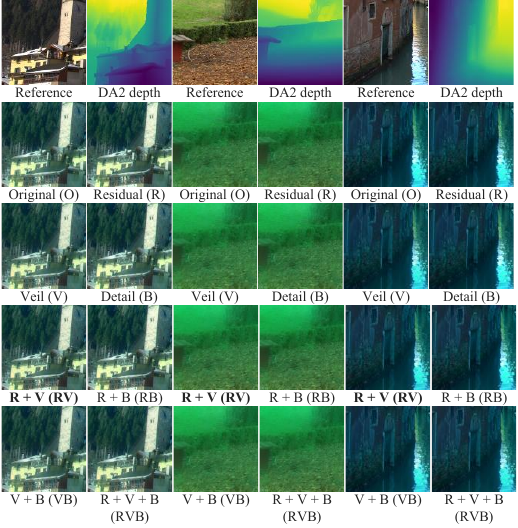}
    \caption{Visual comparison of the original synthesis and depth modulation variants across 3 representative scenes.
    Each 2 column group shows the reference image, DA2 depth map, and eight synthesis variants: O, R, V, B, RV, RB, VB, and RVB.}
    \label{fig:depth_modulation}
\end{figure}

The combination RV addresses these effects jointly.
Relative to O, RV brings the chroma correlation close to zero and reduces SWD from 0.2044 to 0.1885, while maintaining comparable neighborhood precision and structural correspondence.
The broader comparison shows that greater depth attenuation is not sufficient on its own to obtain a preferable supervision set.
In particular, RVB achieves a slightly lower SWD than RV, but the additional detail attenuation also decreases edge similarity with the original reference.
RV is therefore retained as the modulation variant for the downstream study because it offers a balance between the observed depth trends, appearance distribution, and pair structure.
Fig.~\ref{fig:depth_modulation} compares representative outputs from O and the modulation variants.
The subsequent restoration experiments examine whether the differences observed in the synthetic data are reflected in model performance.

%% file: sections/05_downstream.tex
\section{Downstream Restoration Evaluation}
\label{sec:downstream_evaluation}

To evaluate the effectiveness of synthetic supervision, 6 restoration models are trained using 3 public datasets and 2 synthetic datasets constructed by the proposed framework, and evaluated on 3 held-out underwater test sets.
The restoration results are evaluated through objective metrics, visual comparisons, and a human perceptual study.
The experiments further investigate the correspondence between synthetic degradation and learned restoration, as well as the effect of depth-aware modulation on restoration performance.

\subsection{Experimental Setup}
\label{sec:downstream_setup}

\subsubsection{Model Selection}
6 representative underwater image restoration and enhancement methods are selected.
UWCNN~\cite{li2020underwater} and UIEC$^2$-Net~\cite{wang2021uiec} employ convolutional networks, while WaterNet~\cite{li2019underwater} combines multiple enhanced inputs through learned fusion.
U-shape~\cite{peng2023u} uses a Transformer architecture, and DMunderwater~\cite{tang2023underwater} and WF-Diff~\cite{zhao2024wavelet} represent diffusion-based restoration methods.
This selection enables the effects of different training data sources to be examined across diverse restoration architectures.

\subsubsection{Training and Testing Data}
The training data comprise 3 public datasets, UIEB~\cite{li2019underwater}, LSUI~\cite{peng2023u}, and EUVP~\cite{islam2020fast}, along with 2 synthetic datasets, UWNature and UWImgNetSD.
To maintain comparable training scales, the training subsets are selected to approximately match the size of UIEB.
Each restoration model follows its respective default training configuration, with the same settings maintained across different training data sources.
For UWNature, both the original synthesis (O) and its depth-modulated version (RV) are evaluated to examine the effect of depth-aware modulation.
The remaining images from LSUI and EUVP~\cite{islam2020fast} that are not used for training form the combined LSUI\&EUVP test set, which is evaluated alongside U45~\cite{li2019fusion} and RUIE~\cite{liu2020real}.
For consistent testing and evaluation, all images are resized to $256\times256$ pixels.

\subsubsection{Evaluation Metrics}
7 no-reference image quality assessment metrics are adopted to evaluate the restoration results.
PAUQA~\cite{11017762}, MUSIQ~\cite{9710973}, TOPIQ-NR~\cite{10478301}, LIQE~\cite{10204315}, and CLIP-IQA+~\cite{10.1609/aaai.v37i2.25353} provide learned perceptual quality assessments, while UIQM~\cite{7305804} and UCIQE~\cite{7300447} evaluate underwater image quality based on visual attributes.
For the held-out LSUI\&EUVP pairs, PSNR, SSIM\cite{1284395}, and LPIPS~\cite{8578166} are additionally calculated against the provided reference images.
A mean opinion score (MOS) study further assesses the overall visual quality of restored images through human ratings.

\subsection{Qualitative Analysis}
\label{sec:downstream_qualitative}

The qualitative results show that models trained on the proposed synthetic datasets exhibit advantages in several aspects of underwater image restoration and enhancement, including color restoration consistency, generalization, and robustness.

\begin{figure}[!h]
\centering
\includegraphics[width=0.48\textwidth]{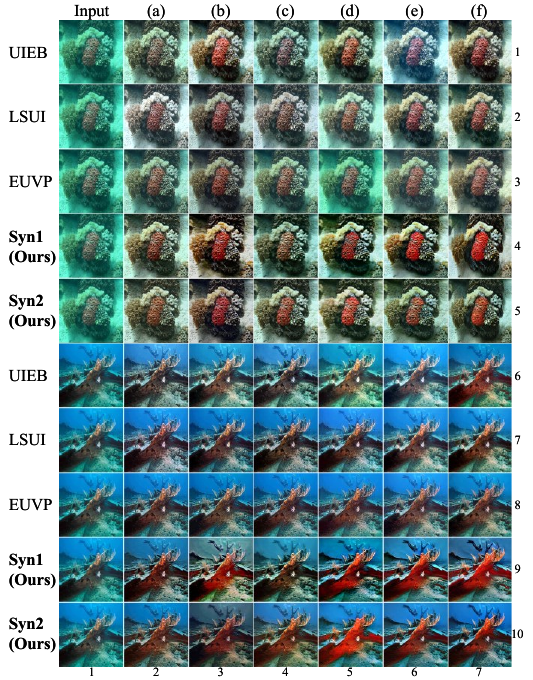}
\caption{Qualitative comparison of restoration and enhancement results on the U45~\cite{li2019fusion} test set.
Models trained on the proposed synthetic datasets (Syn1: UWImgNetSD; Syn2: UWNature; 4th, 5th, 9th, and 10th rows) show more vivid colors and visually consistent restoration than several models trained on public datasets.
Columns (a)--(f) correspond to UWCNN~\cite{li2020underwater}, UIEC$^2$-Net~\cite{wang2021uiec}, WaterNet~\cite{li2019underwater}, U-shape~\cite{peng2023u}, DMunderwater~\cite{tang2023underwater}, and WF-Diff~\cite{zhao2024wavelet}, respectively.
The same conventions apply to Figs.~\ref{fig:downstream_ruie} and~\ref{fig:downstream_lsuieuvp}.}
\label{fig:downstream_u45}
\end{figure}

\subsubsection{Color Restoration and Consistency}
\begin{figure}[!h]
\centering
\includegraphics[width=0.48\textwidth]{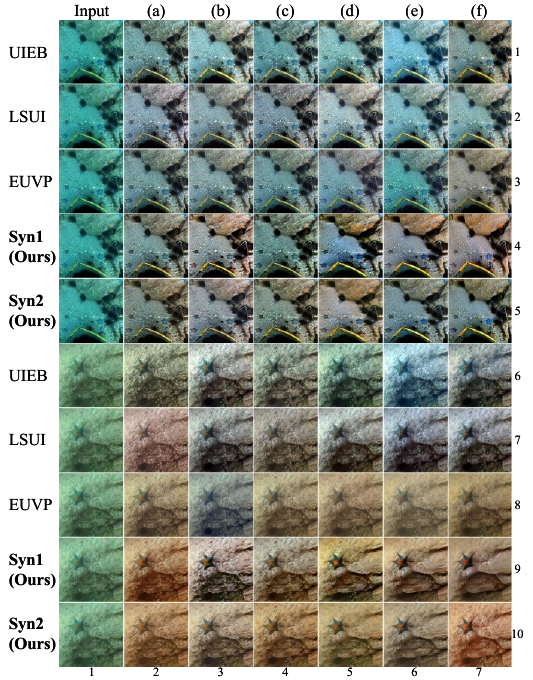}
\caption{Qualitative restoration results on the RUIE~\cite{liu2020real} test set.
Models trained on the proposed synthetic datasets (4th, 5th, 9th, and 10th rows) correct underwater color casts and recover scene details across different images.
The model and training-source layout follows Fig.~\ref{fig:downstream_u45}.}
\label{fig:downstream_ruie}
\end{figure}

As shown in the 4th and 5th rows of Fig.~\ref{fig:downstream_u45}, models trained on the proposed synthetic datasets produce enhanced results on the U45~\cite{li2019fusion} test set with more vivid colors and natural visual appearance.
This improvement is visible across different network architectures, including CNN-based UWCNN, Transformer-based U-shape, and diffusion-based DMunderwater.
Notably, the color correction extends to background regions rather than being limited to the main subject.
For instance, U-shape trained on UWNature recovers coral colors while correcting green and blue color casts in the background water in Fig.~\ref{fig:downstream_u45}.
In contrast, the corresponding U-shape result trained on UIEB~\cite{li2019underwater} mainly enhances the foreground, leaving a noticeable color cast in the background.
This difference is consistent with the training references: UIEB uses selected enhancement results, while the proposed synthetic datasets retain clear in-air images as targets for learning color correction across the entire scene.
This effect is also visible in the RUIE~\cite{liu2020real} results (4th, 5th, 9th, and 10th rows of Fig.~\ref{fig:downstream_ruie}), where the restored images show improved color appearance and clearer details.

\subsubsection{Generalization Capability}
\begin{figure}[!h]
\centering
\includegraphics[width=0.48\textwidth]{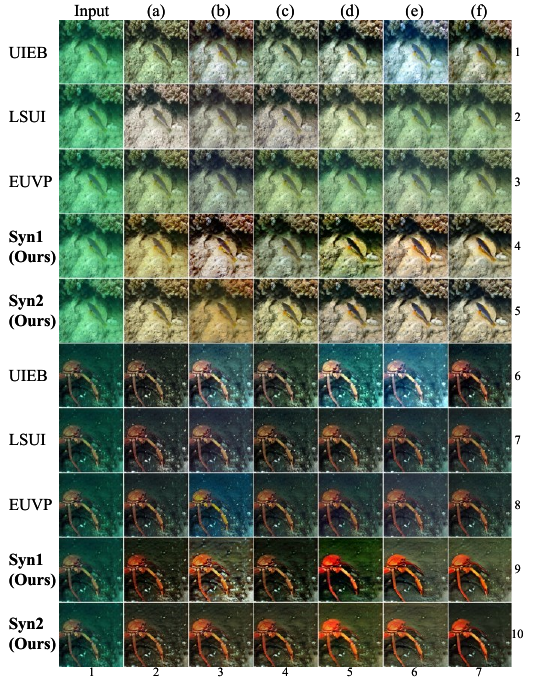}
\caption{Qualitative comparison on the LSUI\&EUVP test set.
Models trained on the proposed synthetic datasets (4th, 5th, 9th, and 10th rows) show clear textures and improved color appearance across different underwater scenes.
The model and training-source layout follows Fig.~\ref{fig:downstream_u45}.}
\label{fig:downstream_lsuieuvp}
\end{figure}

As seen in the 4th, 5th, 9th, and 10th rows of Figs.~\ref{fig:downstream_u45}, \ref{fig:downstream_ruie}, and~\ref{fig:downstream_lsuieuvp}, models trained on the proposed synthetic datasets show visually competitive restoration across U45, RUIE, and LSUI\&EUVP.
For example, in the upper group of Fig.~\ref{fig:downstream_lsuieuvp} (4th and 5th rows), the restored images exhibit clear textures and rich colors, with the original underwater color casts and blurring substantially reduced.
In contrast, some models trained on other datasets, such as EUVP~\cite{islam2020fast}, retain blurring or color distortion in the corresponding examples in Fig.~\ref{fig:downstream_lsuieuvp}.

The benefits also appear across architectures with different modeling capacities.
For WF-Diff, the synthetic supervision preserves fine details and improves color appearance, while simpler networks show fewer residual color casts and better scene visibility in several examples.

\subsection{Quantitative Analysis}
\label{sec:downstream_quantitative}

Tables~\ref{tab:nriqa_learned} and~\ref{tab:nriqa_underwater} compare restoration quality across 6 architectures and 3 test sets using 5 learned and 2 handcrafted no-reference metrics.
The 5 training sources are evaluated alongside the depth-modulated UWNature variant (RV), with all scores reported separately for each architecture and test set.
Compared with the public training datasets, UWNature and UWImgNetSD generally produce restorations with more coherent colors and clearer scene details, which is reflected in their strong learned perceptual scores.

\begin{table*}[t]
\centering
\caption{No-reference image quality comparison using five learned metrics.}
\label{tab:nriqa_learned}
\setlength{\tabcolsep}{1.6pt}
\renewcommand{\arraystretch}{0.98}
\scriptsize
\resizebox{0.95\textwidth}{!}{%
\begin{tabular}{llcclcclcclcclccl}
\toprule
\multirow[c]{3}{*}{Model} &
\multirow[c]{3}{*}{Training Set} &
\multicolumn{3}{c}{PAUQA $\uparrow$} &
\multicolumn{3}{c}{MUSIQ $\uparrow$} &
\multicolumn{3}{c}{TOPIQ-NR $\uparrow$} &
\multicolumn{3}{c}{LIQE $\uparrow$} &
\multicolumn{3}{c}{CLIP-IQA+ $\uparrow$} \\
\cmidrule(lr){3-5}\cmidrule(lr){6-8}\cmidrule(lr){9-11}\cmidrule(lr){12-14}\cmidrule(lr){15-17}
& &
\multirow[c]{2}{*}{U45} &
\multirow[c]{2}{*}{RUIE} &
LSUI &
\multirow[c]{2}{*}{U45} &
\multirow[c]{2}{*}{RUIE} &
LSUI &
\multirow[c]{2}{*}{U45} &
\multirow[c]{2}{*}{RUIE} &
LSUI &
\multirow[c]{2}{*}{U45} &
\multirow[c]{2}{*}{RUIE} &
LSUI &
\multirow[c]{2}{*}{U45} &
\multirow[c]{2}{*}{RUIE} &
LSUI \\
& & & & \&EUVP &
& & \&EUVP &
& & \&EUVP &
& & \&EUVP &
& & \&EUVP \\
\midrule
\multirow{6}{*}{UWCNN~\cite{li2020underwater}} & UIEB~\cite{li2019underwater} & \textbf{41.68} & 35.03 & 42.57 & 48.98 & 41.02 & 46.11 & 0.514 & 0.386 & 0.441 & 2.255 & 1.930 & 2.411 & 0.518 & 0.441 & 0.519 \\
 & LSUI~\cite{peng2023u} & 34.33 & \textbf{37.38} & \textbf{43.77} & 49.78 & 40.49 & 48.88 & 0.478 & 0.385 & 0.467 & 2.205 & 1.999 & 2.707 & 0.515 & \textbf{0.480} & 0.577 \\
 & EUVP~\cite{islam2020fast} & 33.71 & 33.68 & 42.72 & 47.64 & 42.78 & 49.09 & 0.448 & 0.406 & 0.477 & 2.022 & 1.887 & 2.766 & 0.481 & 0.448 & 0.596 \\
 & UWImgNetSD & 30.87 & 32.47 & 40.43 & 50.82 & 41.89 & 50.27 & 0.473 & 0.414 & 0.478 & 2.372 & 2.150 & 2.809 & 0.507 & 0.461 & 0.599 \\
 & UWNature (O) & 34.44 & 34.88 & 42.40 & \textbf{54.36} & \textbf{46.30} & \textbf{52.36} & \textbf{0.516} & \textbf{0.426} & \textbf{0.508} & \textbf{2.628} & \textbf{2.436} & \textbf{3.072} & \textbf{0.545} & 0.473 & \textbf{0.601} \\
 & UWNature (RV) & 33.79 & 32.37 & 39.86 & 46.97 & 38.32 & 47.39 & 0.471 & 0.374 & 0.448 & 2.020 & 1.923 & 2.464 & 0.458 & 0.390 & 0.490 \\
\midrule
\multirow{6}{*}{UIEC$^2$-Net~\cite{wang2021uiec}} & UIEB~\cite{li2019underwater} & 43.60 & 37.34 & 44.25 & 48.51 & 33.70 & 46.09 & 0.549 & 0.381 & 0.458 & 2.623 & 2.038 & 2.540 & 0.549 & 0.419 & 0.515 \\
 & LSUI~\cite{peng2023u} & 40.62 & 38.05 & 44.44 & 50.32 & 38.00 & 50.48 & 0.519 & 0.399 & 0.507 & 2.616 & 2.114 & 2.881 & 0.560 & 0.482 & 0.578 \\
 & EUVP~\cite{islam2020fast} & 40.86 & 38.64 & 44.84 & 50.24 & 40.93 & 50.97 & 0.516 & 0.428 & 0.507 & 2.426 & 2.118 & 2.909 & 0.513 & 0.453 & 0.576 \\
 & UWImgNetSD & \textbf{43.73} & \textbf{41.75} & \textbf{47.42} & \textbf{54.29} & \textbf{47.73} & \textbf{53.12} & 0.565 & \textbf{0.481} & \textbf{0.535} & \textbf{2.975} & \textbf{2.797} & 2.914 & \textbf{0.586} & \textbf{0.558} & \textbf{0.592} \\
 & UWNature (O) & 39.83 & 38.86 & 44.21 & 53.65 & 44.32 & 51.58 & 0.561 & 0.451 & 0.521 & 2.917 & 2.557 & 2.947 & 0.563 & 0.505 & 0.587 \\
 & UWNature (RV) & 38.89 & 38.68 & 43.98 & 54.14 & 42.94 & 51.68 & \textbf{0.576} & 0.448 & 0.523 & 2.761 & 2.381 & \textbf{2.979} & 0.520 & 0.461 & 0.544 \\
\midrule
\multirow{6}{*}{WaterNet~\cite{li2019underwater}} & UIEB~\cite{li2019underwater} & \textbf{41.91} & 35.95 & \textbf{43.09} & 48.00 & 40.13 & 46.81 & 0.516 & 0.379 & 0.446 & 2.299 & 1.949 & 2.538 & 0.522 & 0.460 & 0.544 \\
 & LSUI~\cite{peng2023u} & 36.01 & 34.14 & 42.33 & 49.15 & 38.87 & 47.65 & 0.463 & 0.379 & 0.464 & 2.258 & 2.044 & 2.658 & 0.527 & 0.453 & 0.573 \\
 & EUVP~\cite{islam2020fast} & 32.61 & 33.37 & 42.63 & 47.86 & 42.21 & 48.45 & 0.445 & 0.398 & 0.466 & 2.000 & 1.832 & 2.707 & 0.478 & 0.439 & 0.579 \\
 & UWImgNetSD & 33.73 & 34.02 & 40.79 & 48.31 & 41.07 & 48.27 & 0.443 & 0.374 & 0.449 & 2.213 & 1.968 & 2.540 & 0.486 & 0.435 & 0.572 \\
 & UWNature (O) & 36.14 & \textbf{36.87} & 42.24 & \textbf{54.99} & \textbf{47.23} & \textbf{53.73} & \textbf{0.530} & \textbf{0.446} & \textbf{0.525} & \textbf{2.804} & \textbf{2.508} & \textbf{3.117} & \textbf{0.564} & \textbf{0.521} & \textbf{0.620} \\
 & UWNature (RV) & 35.92 & 35.61 & 42.30 & 49.57 & 39.01 & 47.45 & 0.479 & 0.422 & 0.465 & 2.398 & 2.038 & 2.567 & 0.501 & 0.416 & 0.516 \\
\midrule
\multirow{6}{*}{U-shape~\cite{peng2023u}} & UIEB~\cite{li2019underwater} & \textbf{39.84} & 35.71 & 43.11 & 38.16 & 34.13 & 41.06 & 0.372 & 0.330 & 0.401 & 1.851 & 1.620 & 2.110 & 0.395 & 0.372 & 0.459 \\
 & LSUI~\cite{peng2023u} & 35.83 & 35.65 & 41.74 & 34.98 & 34.77 & 43.44 & 0.365 & 0.367 & 0.429 & 1.597 & 1.730 & 2.334 & 0.383 & 0.406 & 0.522 \\
 & EUVP~\cite{islam2020fast} & 30.16 & 32.96 & 41.60 & 37.81 & 38.44 & 44.39 & 0.376 & 0.368 & 0.434 & 1.474 & 1.694 & 2.351 & 0.359 & 0.377 & 0.498 \\
 & UWImgNetSD & 37.02 & 36.81 & 41.72 & 39.66 & 39.93 & 46.45 & 0.388 & 0.382 & 0.446 & 1.976 & 2.142 & 2.414 & 0.405 & \textbf{0.449} & 0.515 \\
 & UWNature (O) & 36.62 & 38.06 & 43.01 & 39.54 & 38.39 & 45.76 & 0.375 & 0.376 & 0.448 & 1.929 & 2.043 & 2.524 & 0.415 & 0.414 & \textbf{0.526} \\
 & UWNature (RV) & 39.54 & \textbf{39.54} & \textbf{43.89} & \textbf{46.84} & \textbf{42.02} & \textbf{50.76} & \textbf{0.461} & \textbf{0.412} & \textbf{0.492} & \textbf{2.381} & \textbf{2.348} & \textbf{2.776} & \textbf{0.466} & 0.426 & 0.515 \\
\midrule
\multirow{6}{*}{DMunderwater~\cite{tang2023underwater}} & UIEB~\cite{li2019underwater} & \textbf{43.53} & 38.00 & 43.17 & 46.41 & 33.87 & 44.30 & 0.535 & 0.382 & 0.446 & 2.484 & 1.902 & 2.338 & 0.561 & 0.462 & 0.535 \\
 & LSUI~\cite{peng2023u} & 40.21 & 38.18 & 44.34 & 50.59 & 37.99 & 55.17 & 0.521 & 0.404 & 0.581 & 2.542 & 2.135 & 3.256 & 0.579 & 0.487 & 0.616 \\
 & EUVP~\cite{islam2020fast} & 34.36 & 35.05 & 43.86 & 54.32 & 46.65 & 53.83 & 0.510 & 0.430 & 0.553 & 2.271 & 1.861 & 3.013 & 0.542 & 0.505 & 0.628 \\
 & UWImgNetSD & 41.03 & \textbf{41.79} & \textbf{45.60} & \textbf{60.69} & \textbf{56.06} & \textbf{60.06} & \textbf{0.638} & \textbf{0.599} & \textbf{0.644} & \textbf{3.488} & \textbf{3.762} & \textbf{3.968} & \textbf{0.651} & \textbf{0.631} & \textbf{0.693} \\
 & UWNature (O) & 41.00 & 40.92 & 45.48 & 56.38 & 47.13 & 55.53 & 0.606 & 0.510 & 0.592 & 3.002 & 2.864 & 3.602 & 0.610 & 0.524 & 0.632 \\
 & UWNature (RV) & 38.65 & 37.90 & 42.16 & 56.07 & 47.86 & 56.01 & 0.605 & 0.517 & 0.592 & 3.027 & 2.971 & 3.578 & 0.618 & 0.545 & 0.632 \\
\midrule
\multirow{6}{*}{WF-Diff\cite{zhao2024wavelet}} & UIEB~\cite{li2019underwater} & \textbf{42.11} & 34.56 & 41.67 & 48.78 & 36.17 & 45.22 & \textbf{0.542} & 0.379 & 0.448 & 2.683 & 2.050 & 2.486 & 0.552 & 0.437 & 0.499 \\
 & LSUI~\cite{peng2023u} & 39.02 & 37.64 & 43.62 & 49.74 & 37.14 & 51.54 & 0.527 & 0.394 & 0.527 & 2.516 & 2.118 & 3.007 & \textbf{0.562} & 0.468 & 0.592 \\
 & EUVP~\cite{islam2020fast} & 33.77 & 32.71 & 42.98 & 49.34 & 41.11 & 50.69 & 0.499 & 0.402 & 0.515 & 2.221 & 1.759 & 2.914 & 0.513 & 0.466 & \textbf{0.596} \\
 & UWImgNetSD & 38.18 & 39.14 & \textbf{45.05} & \textbf{54.10} & \textbf{48.80} & \textbf{54.57} & 0.524 & \textbf{0.489} & \textbf{0.571} & \textbf{2.732} & \textbf{2.948} & \textbf{3.314} & 0.524 & \textbf{0.512} & 0.572 \\
 & UWNature (O) & 38.85 & \textbf{39.35} & 44.67 & 52.11 & 45.01 & 52.58 & 0.528 & 0.442 & 0.541 & 2.699 & 2.497 & 3.209 & 0.509 & 0.473 & 0.569 \\
 & UWNature (RV) & 40.31 & 37.43 & 43.07 & 51.11 & 44.46 & 51.73 & 0.515 & 0.452 & 0.515 & 2.514 & 2.539 & 3.039 & 0.466 & 0.422 & 0.506 \\
\bottomrule
\end{tabular}%
}
\end{table*}

\begin{table}[t]
\centering
\caption{Underwater image quality comparison using UIQM and UCIQE.}
\label{tab:nriqa_underwater}
\setlength{\tabcolsep}{1.2pt}
\renewcommand{\arraystretch}{0.98}
\scriptsize
\resizebox{0.95\columnwidth}{!}{%
\begin{tabular}{llcclccl}

\toprule
\multirow[c]{3}{*}{Model} &
\multirow[c]{3}{*}{Training Set} &
\multicolumn{3}{c}{UIQM $\uparrow$} &
\multicolumn{3}{c}{UCIQE $\uparrow$} \\
\cmidrule(lr){3-5}\cmidrule(lr){6-8}
& &
\multirow[c]{2}{*}{U45} &
\multirow[c]{2}{*}{RUIE} &
LSUI &
\multirow[c]{2}{*}{U45} &
\multirow[c]{2}{*}{RUIE} &
LSUI \\
& & & & \&EUVP & & & \&EUVP \\
\midrule
\multirow{6}{*}{UWCNN~\cite{li2020underwater}} & UIEB~\cite{li2019underwater} & \textbf{0.688} & 0.347 & 0.795 & 0.5507 & 0.5115 & 0.5732 \\
 & LSUI~\cite{peng2023u} & 0.646 & \textbf{0.746} & \textbf{0.980} & 0.5390 & 0.5113 & 0.5773 \\
 & EUVP~\cite{islam2020fast} & 0.454 & 0.334 & 0.746 & 0.5369 & 0.5177 & 0.5745 \\
 & UWImgNetSD & 0.639 & 0.547 & 0.913 & \textbf{0.5612} & \textbf{0.5493} & \textbf{0.5979} \\
 & UWNature (O) & 0.613 & 0.608 & 0.882 & 0.5485 & 0.5213 & 0.5798 \\
 & UWNature (RV) & 0.565 & 0.501 & 0.848 & 0.5421 & 0.5266 & 0.5748 \\
\midrule
\multirow{6}{*}{UIEC$^2$-Net~\cite{wang2021uiec}} & UIEB~\cite{li2019underwater} & 0.797 & 0.823 & 0.937 & \textbf{0.6172} & 0.5726 & 0.6121 \\
 & LSUI~\cite{peng2023u} & 0.655 & 0.773 & 0.866 & 0.5785 & 0.5554 & 0.5835 \\
 & EUVP~\cite{islam2020fast} & 0.471 & 0.289 & 0.761 & 0.5302 & 0.4960 & 0.5745 \\
 & UWImgNetSD & \textbf{1.101} & \textbf{0.950} & \textbf{1.286} & 0.5989 & \textbf{0.5796} & \textbf{0.6310} \\
 & UWNature (O) & 0.737 & 0.799 & 0.853 & 0.5810 & 0.5480 & 0.6035 \\
 & UWNature (RV) & 0.711 & 0.747 & 0.891 & 0.5745 & 0.5377 & 0.6014 \\
\midrule
\multirow{6}{*}{WaterNet~\cite{li2019underwater}} & UIEB~\cite{li2019underwater} & 0.671 & 0.398 & 0.788 & 0.5480 & 0.5154 & 0.5756 \\
 & LSUI~\cite{peng2023u} & 0.584 & 0.642 & \textbf{0.990} & 0.5315 & 0.5116 & 0.5689 \\
 & EUVP~\cite{islam2020fast} & 0.462 & 0.344 & 0.758 & 0.5353 & 0.5181 & 0.5722 \\
 & UWImgNetSD & 0.524 & 0.412 & 0.918 & 0.5549 & 0.5412 & 0.5958 \\
 & UWNature (O) & 0.635 & 0.630 & 0.869 & 0.5730 & 0.5416 & 0.5897 \\
 & UWNature (RV) & \textbf{0.750} & \textbf{0.775} & 0.985 & \textbf{0.5806} & \textbf{0.5656} & \textbf{0.6045} \\
\midrule
\multirow{6}{*}{U-shape~\cite{peng2023u}} & UIEB~\cite{li2019underwater} & 0.614 & 0.569 & 0.825 & \textbf{0.5964} & 0.5529 & 0.6012 \\
 & LSUI~\cite{peng2023u} & 0.668 & \textbf{0.883} & 0.923 & 0.5654 & 0.5489 & 0.5861 \\
 & EUVP~\cite{islam2020fast} & 0.390 & 0.364 & 0.732 & 0.5136 & 0.5159 & 0.5638 \\
 & UWImgNetSD & \textbf{0.676} & 0.815 & \textbf{1.154} & 0.5833 & \textbf{0.5641} & \textbf{0.6250} \\
 & UWNature (O) & 0.627 & 0.717 & 0.850 & 0.5734 & 0.5517 & 0.6017 \\
 & UWNature (RV) & 0.474 & 0.703 & 0.783 & 0.5507 & 0.5379 & 0.5720 \\
\midrule
\multirow{6}{*}{DMunderwater~\cite{tang2023underwater}} & UIEB~\cite{li2019underwater} & 0.625 & 0.557 & 0.714 & 0.5854 & 0.5430 & 0.5802 \\
 & LSUI~\cite{peng2023u} & 0.635 & \textbf{0.774} & 0.855 & 0.5827 & 0.5700 & 0.5853 \\
 & EUVP~\cite{islam2020fast} & 0.514 & 0.375 & 0.753 & 0.5124 & 0.5020 & 0.5650 \\
 & UWImgNetSD & \textbf{0.790} & 0.768 & \textbf{0.972} & \textbf{0.5957} & \textbf{0.5710} & \textbf{0.6131} \\
 & UWNature (O) & 0.641 & 0.702 & 0.833 & 0.5667 & 0.5270 & 0.5887 \\
 & UWNature (RV) & 0.729 & 0.628 & 0.901 & 0.5749 & 0.5364 & 0.5967 \\
\midrule
\multirow{6}{*}{WF-Diff\cite{zhao2024wavelet}} & UIEB~\cite{li2019underwater} & 0.672 & 0.486 & 0.821 & \textbf{0.6026} & 0.5552 & 0.6038 \\
 & LSUI~\cite{peng2023u} & 0.667 & \textbf{0.807} & 0.860 & 0.5786 & \textbf{0.5687} & 0.5869 \\
 & EUVP~\cite{islam2020fast} & 0.484 & 0.304 & 0.714 & 0.5332 & 0.5083 & 0.5725 \\
 & UWImgNetSD & 0.771 & 0.679 & \textbf{0.938} & 0.5931 & 0.5662 & \textbf{0.6120} \\
 & UWNature (O) & 0.677 & 0.614 & 0.823 & 0.5761 & 0.5369 & 0.5998 \\
 & UWNature (RV) & \textbf{0.787} & 0.626 & 0.914 & 0.5818 & 0.5512 & 0.6070 \\
\bottomrule
\end{tabular}%
}
\end{table}

The clearest trend appears in the learned perceptual metrics.
When the 5 original training sources are compared, UWNature and UWImgNetSD account for 77 of the 90 highest scores across PAUQA, MUSIQ, TOPIQ-NR, LIQE, and CLIP-IQA+.
Their advantage is especially consistent under MUSIQ and LIQE, where one of the 2 synthetic datasets ranks first for every architecture on all 3 test sets.
This pattern is shared by convolutional, Transformer-based, and diffusion-based restoration networks, indicating that the benefit of synthetic supervision is not tied to a particular restoration architecture.
For example, UWNature raises WaterNet's MUSIQ score on U45 from $48.00$ with UIEB supervision to $54.99$, while UWImgNetSD produces the highest MUSIQ scores for DMunderwater across all 3 test sets.

These results are closely related to how the training pairs are constructed.
The proposed framework retains a clear in-air image as the restoration target and synthesizes underwater degradation only on the input side, so the same scene content is presented through a degraded input and a coherent clean reference.
This provides a consistent target for correcting color casts across foreground and background regions while recovering natural color relationships and local textures.
MUSIQ evaluates quality at multiple spatial scales, TOPIQ-NR combines semantic guidance with local distortion assessment, and LIQE and CLIP-IQA+ incorporate visual-language representations of image quality.
Their agreement suggests that the synthetic training pairs help the models improve several aspects of visual quality together, rather than simply increasing saturation or contrast.

The rankings also reveal differences between general perceptual assessment and quality prediction tailored to underwater images.
PAUQA learns underwater perceptual quality from chromatic, luminance, and spatial information, and its preferences do not always follow MUSIQ or LIQE.
On U45, UIEB obtains the highest PAUQA score for 5 of the 6 architectures among the original training sources, even though the 2 synthetic datasets consistently lead MUSIQ and LIQE.
On RUIE, UWNature leads PAUQA for WaterNet, U-shape, and WF-Diff, while UWImgNetSD leads for UIEC$^2$-Net and DMunderwater.
PAUQA emphasizes underwater color and illumination, whereas general perceptual metrics also consider the natural appearance and structural detail of the restored scene, so the same color correction may be rewarded differently.
The different preferences across U45 and RUIE also reflect the diversity of real underwater appearances observed in Section~\ref{sec:intrinsic_validation}, where color, luminance, contrast, and detail vary together rather than forming isolated degradation types.

The handcrafted metrics provide a complementary view of these image attributes.
UIQM combines colorfulness, sharpness, and contrast.
UCIQE is determined by chroma variation, luminance contrast, and saturation.
On RUIE, LSUI achieves the highest UIQM score for 5 architectures among the 5 original training sources, whereas UWNature and UWImgNetSD continue to lead MUSIQ, TOPIQ-NR, and LIQE for every architecture.
On LSUI\&EUVP, UWImgNetSD attains the highest UCIQE for all 6 architectures and the highest UIQM for 4 of them among the original training sources.
A high UIQM score can be driven by stronger colorfulness, sharpening, and contrast, even when the overall appearance is less natural; this helps explain why LSUI performs well under UIQM on RUIE while the synthetic datasets lead several learned perceptual metrics.
The synthetic supervision supports both types of improvement, but the balance varies with the training source, network, and test domain.

The results also distinguish the 2 datasets generated by the same framework.
UWNature performs particularly well with UWCNN and WaterNet under several learned metrics, whereas UWImgNetSD provides the strongest perceptual scores for DMunderwater and often for UIEC$^2$-Net.
The 2 datasets use the same construction framework, but UWImgNetSD takes AI-generated images as clean references, while UWNature uses real natural photographs.
AI-generated images often have more vivid colors, higher saturation, and stronger contrast, whereas natural photographs tend to retain more moderate colors and gradual changes in tone and texture.
The vivid references in UWImgNetSD may encourage stronger color and contrast restoration, which is consistent with its high UIQM and UCIQE scores on several test sets.
The difference in reference appearance may also help explain why UWImgNetSD achieves particularly strong perceptual scores with DMunderwater and UIEC$^2$-Net, while UWNature performs well with UWCNN and WaterNet.
Taken together, the comparisons across the 3 test sets support the central role of reliable clean targets and plausible synthetic inputs in producing transferable restoration supervision.

The RV rows show how the same evaluation changes when the synthetic input degradation is modified while the clean reference is retained.
For U-shape on U45, RV increases MUSIQ from $39.54$ to $46.84$ and TOPIQ-NR from $0.375$ to $0.461$, while UIQM decreases from $0.627$ to $0.474$.
The different responses of the other architectures are examined using paired image-level statistics in Section~\ref{sec:depth_modulation_ablation}.

Full-reference metrics, including PSNR, SSIM, and LPIPS, were also computed on the held-out LSUI\&EUVP pairs.
These metrics measure agreement with the benchmark references, whose appearances reflect the enhancement methods and selection procedures used in dataset construction.
Models trained on LSUI or EUVP may therefore be favored by these metrics, as their training targets follow the same reference conventions even when training and test images do not overlap.
In contrast, our framework preserves original clear images as restoration targets.
We therefore emphasize no-reference metrics and human perceptual evaluation to assess restoration quality across different supervision sources without favoring a particular reference style.

\subsection{Human Perceptual Evaluation}
\label{sec:human_perceptual}

A human perceptual study is conducted to assess the overall visual quality of the restored images.
10 participants independently rate outputs from UWCNN, U-shape, and WF-Diff trained on the 5 original datasets, without the RV condition.
The study covers 27 input scenes from the 3 test sets for each architecture, giving 405 unique restoration stimuli.
Each stimulus is evaluated by 2 participants on a five-point scale, resulting in 810 main ratings and 70 repeated ratings for quality control.
The image order is randomized independently for each participant, and the training source and restoration model are not disclosed.
Participants judge the restored images by color naturalness, visibility, detail preservation, and visible artifacts.

\begin{table}[h]
    \centering
    \caption{Mean opinion scores (MOS) and 95\% confidence intervals for different training datasets.}
    \label{tab:mos}
    \footnotesize
    \setlength{\tabcolsep}{2pt}
    \renewcommand{\arraystretch}{1.25}
    \begin{tabular*}{\columnwidth}{
        @{}l@{\extracolsep{\fill}}ccccc@{}
    }
        \toprule
        \raisebox{-.5\height}{\shortstack{Training\\ Data}} &
        \raisebox{-.5\height}{UIEB} &
        \raisebox{-.5\height}{LSUI} &
        \raisebox{-.5\height}{EUVP} &
        \raisebox{-.5\height}{UWNature} &
        \raisebox{-.5\height}{UWImgNetSD} \\
        \midrule
        MOS $\uparrow$ &
        3.321 &
        3.469 &
        3.284 &
        3.673 &
        \textbf{3.784} \\
        95\% CI &
        $[3.17, 3.48]$ &
        $[3.29, 3.64]$ &
        $[3.10, 3.47]$ &
        $[3.52, 3.83]$ &
        $[3.59, 3.97]$ \\
        \bottomrule
    \end{tabular*}
\end{table}

As shown in Table~\ref{tab:mos}, UWImgNetSD and UWNature achieve the highest overall MOS scores.
Unlike individual quality metrics, MOS reflects the combined human perception of color naturalness, visibility, detail preservation, and artifacts.
Across 81 matched architecture--scene groups, UWNature exceeds UIEB by $0.352$ (95\% CI $[0.160,0.549]$) and EUVP~\cite{islam2020fast} by $0.389$ (95\% CI $[0.216,0.562]$).
Its margin over LSUI is smaller at $0.204$ (95\% CI $[0.012,0.401]$), with greater sensitivity to participant variation.
The difference between UWNature and UWImgNetSD is $-0.111$ (95\% CI $[-0.303,0.086]$).
The favorable ratings suggest that retaining clear reference images while synthesizing degradation on the input side supports visually coherent restoration across different scenes and architectures.
UWImgNetSD's slightly higher mean score may be associated with the vivid colors and stronger contrast of its AI-generated references, whereas UWNature uses natural photographs with more moderate colors and tonal variations.
Together with the objective results in Section~\ref{sec:downstream_quantitative}, the subjective evaluation supports the perceptual quality of restoration learned from the proposed synthetic supervision.

\subsection{Degradation and Restoration Reciprocity Analysis}
\label{sec:restoration_reciprocity}

To investigate what restoration models learn from synthetic supervision, the degradation introduced during data generation is compared with the corrections produced on real underwater images.
Using the five-dimensional appearance descriptor $\mathbf{f}(\cdot)$ defined in Eq.~\eqref{eq:appearance_descriptor}, synthetic degradation is represented by $\mathbf{d}_j=\mathbf{f}(x_j)-\mathbf{f}(y_j)$, while the restoration of a real underwater image $u$ is represented by $\mathbf{r}(u)=\mathbf{f}(H_{\theta}(u))-\mathbf{f}(u)$.
The analysis uses U-shape trained on the original UWNature dataset and evaluates its outputs on U45, RUIE, and LSUI\&EUVP.
For each real input, the 5 nearest synthetic degraded samples are selected in the standardized descriptor space, and their average inverse degradation direction is
\begin{equation}
    \widehat{\mathbf{r}}_K(u)
    =-\frac{1}{K}\sum_{j\in\mathcal{N}_K(u)}\mathbf{d}_j,
    \qquad K=5,
    \label{eq:reciprocity_direction}
\end{equation}
where $\mathcal{N}_K(u)$ denotes the indices of the selected synthetic samples.
Cosine similarity between $\widehat{\mathbf{r}}_K(u)$ and $\mathbf{r}(u)$ measures their directional agreement, with global and randomized estimates used for comparison.

The results show that restoration corrections generally oppose synthetic degradation in color, contrast, and detail, with less consistent changes in luminance.
The mean directional similarity reaches $0.910$ using the nearest synthetic samples, compared with $0.861$ using the global mean direction and lower values under randomized comparisons.
The improvement over the global estimate is $0.049$, with a 95\% confidence interval of $[0.040,0.058]$.

These findings provide insight into how the proposed synthetic pairs guide restoration learning.
By retaining clear natural images as references and generating underwater degradation on the input side, the framework establishes a consistent relationship between degraded appearances and their corresponding restoration targets.
Through training, the model learns to reverse characteristic underwater changes in color, contrast, and detail, allowing the acquired restoration behavior to transfer to real underwater images.
The strong global directional agreement reflects common correction patterns across underwater scenes, while the additional improvement from matching similar synthetic degradations suggests that the learned corrections also respond to differences in input appearance.
Therefore, the proposed framework provides not only visually plausible synthetic training pairs but also meaningful restoration supervision whose learned correction directions correspond to those applied to real underwater images.

\subsection{Effect of Depth Modulation on Restoration}
\label{sec:depth_modulation_ablation}

\begin{figure}[t]
\centering
\includegraphics[width=0.48\textwidth]{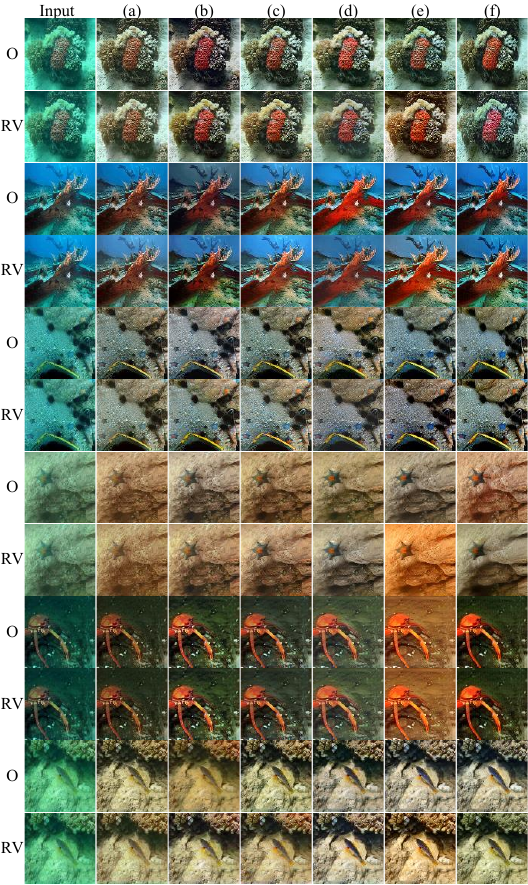}
\caption{Qualitative comparison of restoration results using the original UWNature synthesis ($O$) and its depth-aware modulation ($RV$) on U45, RUIE, and LSUI\&EUVP.
Each test set contains the same input image restored by models trained with $O$ and $RV$.
Columns (a)--(f) correspond to UWCNN, UIEC$^2$-Net, WaterNet, U-shape, DMunderwater, and WF-Diff, respectively.}
\label{fig:rv_restoration}
\end{figure}

To investigate the influence of depth-aware degradation on restoration learning, the original UWNature synthesis ($O$) is compared with its depth-modulated variant ($RV$).
The modulation redistributes degradation residuals and introduces depth-dependent veiling into the synthetic inputs while preserving their clean references.
All 6 restoration architectures are trained separately using $O$ and $RV$ under identical settings for each architecture and evaluated on the same U45, RUIE, and LSUI\&EUVP test inputs.
Fig.~\ref{fig:rv_restoration} presents representative restoration results under the 2 training conditions.
The quantitative comparison uses the 7 no-reference metrics in Tables~\ref{tab:nriqa_learned} and~\ref{tab:nriqa_underwater}, together with PSNR, SSIM, and LPIPS on LSUI\&EUVP.
Differences are assessed using 20,000 paired bootstrap resamples of the image-level scores, with positive differences defined to favor $RV$ for every metric, including LPIPS.

The results show that depth-aware modulation produces different effects across restoration architectures and evaluation metrics.
Based on the 95\% paired bootstrap confidence intervals, $RV$ outperforms $O$ in 51 of the 144 comparisons, underperforms in 75, and shows no clear difference in the remaining 18.
U-shape benefits most frequently, with improvements in 17 of its 24 comparisons, whereas UWCNN improves in only 3.
The metric responses also differ within the same architecture.
For example, U-shape trained with $RV$ achieves higher MUSIQ and TOPIQ-NR scores on U45, while its UIQM score decreases.
These results indicate that modifying the spatial distribution of synthetic degradation can improve certain aspects of restoration without necessarily improving all visual attributes.

This behavior reveals an important relationship between degradation modeling and restoration supervision.
Depth-aware modulation improves the depth-dependent spatial behavior of synthetic degradation while changing the distribution and complexity of the degraded inputs encountered during training.
Since the clean references remain unchanged, restoration models must learn to recover the same target appearances from different degradation patterns.
Architectures capable of integrating information across spatial regions, such as U-shape, may benefit more from the additional depth-dependent variation, whereas other networks may learn more effectively from the original degradation distribution.
The results therefore suggest that effective synthetic supervision depends on both the plausibility of the generated degradation and the ability of restoration architectures to learn transferable inverse mappings.

\subsection{Real-world Validation}
\label{sec:real_world_validation}

\begin{figure*}[t]
    \centering
    \includegraphics[width=0.98\textwidth]{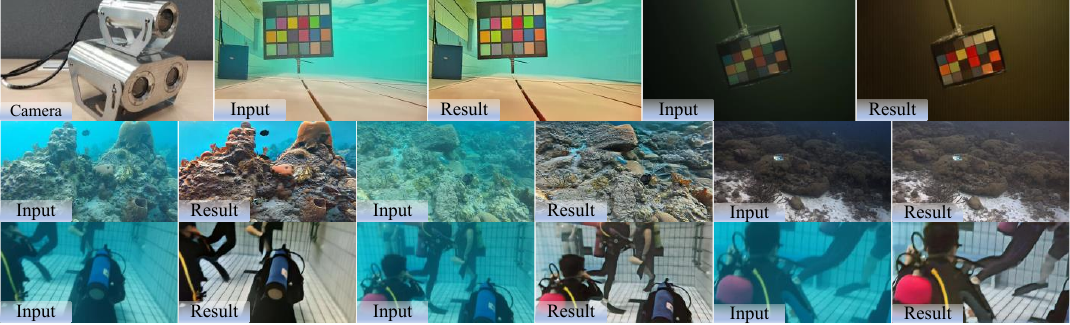}
    \caption{Real-world underwater restoration examples using DMunderwater trained on UWNature.
            The figure presents the camera setup and restoration results for color card, coral reefs, and divers in a swimming pool.}
    \label{fig_app}
\end{figure*}

To examine restoration in real underwater settings, we evaluated color card images captured with a self-developed camera under blue cast and low light conditions, together with images from natural underwater scenes and a swimming pool.
DMunderwater~\cite{tang2023underwater}, trained on UWNature, was applied to these images.
Fig.~\ref{fig_app} presents representative restoration results from natural underwater environments and swimming pool scenes.
The restored images show reduced color casts, clearer scene details, and improved visibility across different underwater conditions.
The results of color card experiments also show improved color appearance and clearer boundaries between patches.
These examples illustrate the qualitative restoration performance of models trained with the proposed synthetic supervision in real underwater settings.

%% file: sections/06_conclusion.tex
\section{Conclusion}

This paper presented a generative data framework for constructing reliable paired supervision for underwater image restoration and enhancement.
By preserving clear images as references and synthesizing underwater-degraded inputs, the framework provides consistent restoration targets without requiring paired underwater observations.
Using UWNature as the main instance, we evaluated the generated supervision through appearance distribution, structural consistency, and spatial degradation characteristics across relative scene depth.
Restoration experiments across 6 architectures and 3 test sets, together with human perceptual evaluation, demonstrated the effectiveness of the synthetic supervision on real underwater images.
The degradation and restoration reciprocity analysis further showed that synthetic pairs encode restoration directions that can be learned and transferred to real underwater images.
The depth modulation study showed that improving the spatial plausibility of synthetic degradation does not necessarily improve restoration performance across architectures and metrics.

The significance of this work extends beyond the construction of individual synthetic datasets.
The same framework also generates UWImgNetSD from AI-generated reference images, demonstrating its applicability to different clear image sources.
Our study highlights the fundamental role of training data in shaping the restoration behavior learned by different networks.
Reliable supervision combines clear reference targets, plausible degraded inputs, and consistent relationships between them.
The quality of synthetic supervision is reflected in both the intrinsic properties of the generated pairs and their effectiveness in downstream restoration.
By emphasizing the construction and validation of reliable supervision, this work provides a scalable data-centric approach to underwater image restoration and offers broader insights into learning from synthetic data when reliable paired observations are difficult to obtain.

%% file: appendix/supplement.tex
The analyses below examine how the properties of synthetic image pairs relate to the restoration behavior learned from them.
They extend the intrinsic validation and downstream comparisons in Sections IV and V.

\subsection{Evaluation Details}
The component results are reported for each restoration architecture, training source, and test set, using the image-level scores underlying Table IV.
UIQM follows the xueleichen-compatible implementation, while UCIQE uses the same protocol as the main manuscript.
The five original training sources are compared separately from the depth-modulated UWNature (RV) condition.

\subsection{Appearance and Structural Consistency}
\subsubsection{Appearance statistics}
The descriptor normalization and PCA basis are fitted to 2,850 real underwater images and then applied to the full real collection and synthetic samples.
This provides a common coordinate system for the distribution analysis in Fig.~2.
The correlation between low contrast and low detail in real underwater images is $0.5887$, with a bootstrap 95\% confidence interval of $[0.5605,0.6157]$ after controlling for the source dataset.
This coupling is consistent with the use of representative, overlapping appearance targets for I2IT rather than separate physical degradation categories.

\subsubsection{Structure of paired images}
\begin{figure*}[!t]
\centering
\includegraphics[width=0.98\textwidth]{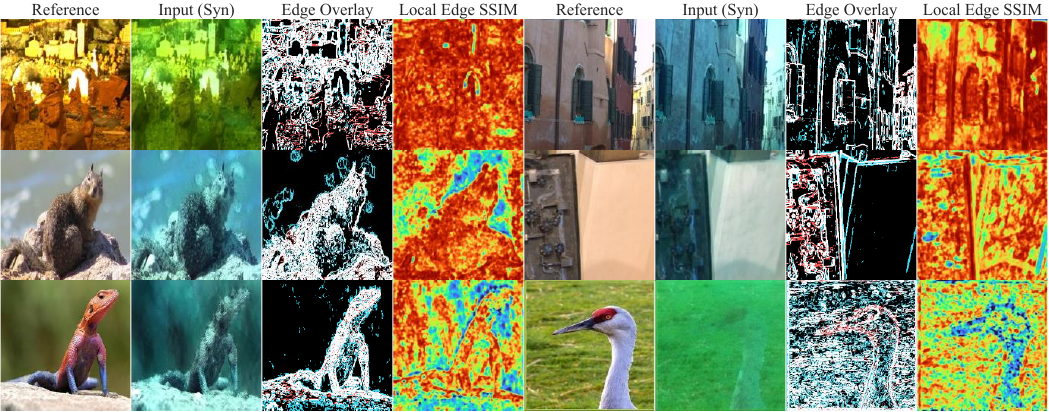}
\caption{Structural consistency in 6 UWNature pairs, ordered from higher to lower edge SSIM with two pairs per row.
Each pair shows the reference, synthetic input, edge overlay (white: shared; red: reference-only; cyan: synthetic-only), and local edge-SSIM map (blue--violet: lower; orange--red: higher similarity).
Main object geometry persists in the upper examples, while local detail loss becomes more prominent in the lower examples.}
\label{fig:sup-pair-cases}
\end{figure*}

Table~\ref{tab:sup-structure} reports the quantiles of grayscale correlation, edge correlation, and edge SSIM for 1,523 UWNature pairs.
The high median correlations indicate that the generated inputs usually retain the scene layout needed for paired training, whereas the lower edge SSIM tail ($P5=0.474$) reveals appreciable variation in local structure preservation.
Phase registration estimates no translational shift for 1,521 pairs and shifts of one and two pixels for the remaining pairs, suggesting that global displacement is not the principal source of this variation.
Fig.~\ref{fig:sup-pair-cases} makes these differences visible in 6 examples spanning higher, intermediate, and lower edge SSIM values.
In the crowd and street scenes (0.855 and 0.837), the underwater color transformation leaves the principal silhouettes, architectural lines, and spatial layout well aligned with their references.
The squirrel and mechanical scenes (0.711 and 0.707) retain recognizable objects but show changes in fine texture and local edge responses that are less apparent in the color images alone.
The lizard remains identifiable at an edge SSIM of 0.571, whereas the bird at 0.530 shows much greater loss of subject visibility, illustrating that a similar range of edge scores can encompass different perceptual consequences.
The edge overlays distinguish shared boundaries from reference-only and newly emphasized edges, and the local similarity maps show where these differences occur.
Taken with the distribution in Table~\ref{tab:sup-structure}, the examples indicate that synthesis largely preserves the image geometry needed to pair a degraded input with its unchanged clear reference, while the amount of recoverable fine structure varies across samples.
The edge SSIM values in Table I and Table II use percentile-normalized and raw Sobel responses, respectively.

\begin{table}[h]
\caption{Quantiles of structural consistency over 1,523 UWNature pairs.}
\label{tab:sup-structure}
\centering\scriptsize\setlength{\tabcolsep}{3.4pt}
\begin{tabular}{lrrrrr}\toprule
Measure & P5 & P25 & Median & P75 & P95 \\
\midrule
Gray correlation & 0.794 & 0.905 & 0.943 & 0.967 & 0.988 \\
Edge correlation & 0.704 & 0.803 & 0.860 & 0.908 & 0.954 \\
Edge SSIM & 0.474 & 0.632 & 0.713 & 0.789 & 0.884 \\
\bottomrule\end{tabular}\end{table}

\subsection{Controls for Spatial Depth Analysis}
\begin{figure}[!t]
\centering
\includegraphics[width=0.48\textwidth]{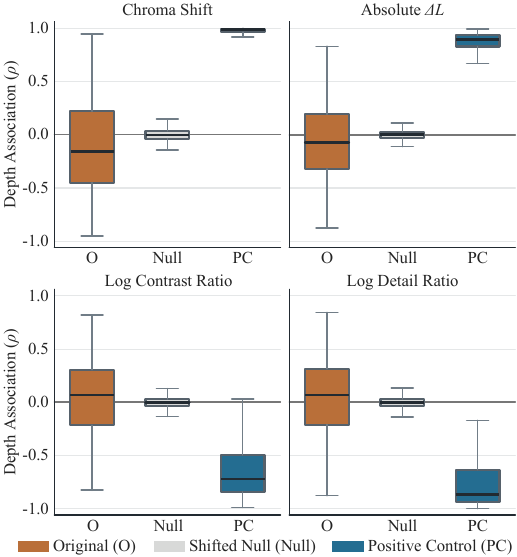}
\caption{Image-level depth associations for four degradation attributes using DA2 relative-depth estimates.
Boxes show the original synthesis (O), the within-image median of 50 shifted-depth comparisons (Null), and the imposed positive control (PC).
The near-zero null distributions and correctly signed control correlations distinguish measured spatial trends from random correspondence.}
\label{fig:sup-depth-null}
\end{figure}

To examine the depth associations in Section IV-C, the estimated relative depth maps are spatially shifted 50 times per image to provide a null comparison.
A positive control introduces a known spatial trend into the degradation attributes.
Table~\ref{tab:sup-depth-control} shows that shifting the depth maps reduces the correlations to nearly zero, while the positive controls recover the directions of the imposed changes.
Fig.~\ref{fig:sup-depth-null} further shows the image-level distributions behind these medians, with each shifted-null observation summarized over the 50 shifts of the same image.
The shifted-depth distributions cluster around zero for all four attributes, in contrast to the broader and systematically displaced original distributions.
The positive controls recover strong positive associations for chroma shift and absolute luminance change and negative associations for the log contrast and detail ratios, following the directions built into the controls.
This separation indicates that the negative chroma association and positive detail-ratio association of the original synthesis in Table I reflect spatial behavior of the generated images rather than a correlation imposed by the block-wise calculation.
In particular, the original depth relationships do not consistently follow the attenuation pattern expected at greater viewing distances, motivating the depth modulation analysis in Section IV-D.
The modulation changes the spatial properties of the synthetic inputs without changing their clear references; its influence on learned restoration is examined separately through the O/RV comparison in Section V-F and below.

\begin{table}[h]
\caption{Depth association controls using DA2 relative depth, reported as median Spearman correlations.}
\label{tab:sup-depth-control}
\centering\scriptsize\setlength{\tabcolsep}{3.1pt}
\begin{tabular}{lrr}\toprule
Attribute & Shifted depth & Positive control \\
\midrule
Chroma shift & 0.0005 & 0.9875 \\
Absolute $\Delta L$ & 0.0011 & 0.8964 \\
Log contrast ratio & $-0.0012$ & $-0.7209$ \\
Log detail ratio & $-0.0020$ & $-0.8635$ \\
\bottomrule
\end{tabular}
\end{table}

\subsection{Additional Restoration Quality Results}
\subsubsection{Full-reference metrics}
Table~\ref{tab:sup-fr} gives the complete PSNR, SSIM, and LPIPS results on LSUI\&EUVP for the 6 restoration architectures and 6 training conditions.
LSUI or EUVP supervision gives the highest PSNR and SSIM for all 6 architectures, reflecting the close agreement of these models with the supplied LSUI\&EUVP benchmark references.
The ranking changes when visual quality rather than reference similarity is considered: for WF-Diff, LSUI yields 23.59 dB PSNR compared with 18.04 dB for UWImgNetSD, while the corresponding architecture-level MOS scores are 3.778 and 4.167, respectively.
These results illustrate the distinction between reproducing an enhancement-based benchmark target and producing visually preferred restoration on real underwater images.
The evaluation in the main manuscript therefore combines full-reference agreement with learned perceptual metrics and human ratings rather than relying on a single reference convention.

\begin{table}[!t]
\centering
\caption{Full-reference restoration quality on LSUI\&EUVP.}
\label{tab:sup-fr}
\setlength{\tabcolsep}{1.2pt}
\renewcommand{\arraystretch}{0.90}
\scriptsize
\resizebox{0.95\columnwidth}{!}{%
\begin{tabular}{llccc}
\toprule
Model & Training Set & PSNR $\uparrow$ & SSIM $\uparrow$ & LPIPS $\downarrow$ \\
\midrule
\multirow{6}{*}{UWCNN~\cite{li2020underwater}} & UIEB~\cite{li2019underwater} & 19.90 & 0.8143 & 0.2760 \\
 & LSUI~\cite{peng2023u} & 19.56 & 0.8229 & 0.2600 \\
 & EUVP~\cite{islam2020fast} & \textbf{21.50} & \textbf{0.8449} & \textbf{0.2418} \\
 & UWImgNetSD & 19.82 & 0.7999 & 0.2770 \\
 & UWNature (O) & 20.30 & 0.8175 & 0.2598 \\
 & UWNature (RV) & 20.46 & 0.8267 & 0.2766 \\
\midrule
\multirow{6}{*}{UIEC$^2$-Net~\cite{wang2021uiec}} & UIEB~\cite{li2019underwater} & 19.34 & 0.8063 & 0.2666 \\
 & LSUI~\cite{peng2023u} & \textbf{21.47} & \textbf{0.8501} & \textbf{0.2215} \\
 & EUVP~\cite{islam2020fast} & 20.73 & 0.8325 & 0.2457 \\
 & UWImgNetSD & 17.20 & 0.7038 & 0.3611 \\
 & UWNature (O) & 19.30 & 0.8028 & 0.2789 \\
 & UWNature (RV) & 19.36 & 0.7881 & 0.2777 \\
\midrule
\multirow{6}{*}{WaterNet~\cite{li2019underwater}} & UIEB~\cite{li2019underwater} & 18.80 & 0.8059 & 0.2824 \\
 & LSUI~\cite{peng2023u} & 20.24 & 0.8214 & 0.2727 \\
 & EUVP~\cite{islam2020fast} & \textbf{21.41} & \textbf{0.8411} & \textbf{0.2542} \\
 & UWImgNetSD & 18.80 & 0.7660 & 0.2979 \\
 & UWNature (O) & 19.80 & 0.8097 & 0.2647 \\
 & UWNature (RV) & 20.00 & 0.7728 & 0.2950 \\
\midrule
\multirow{6}{*}{U-shape~\cite{peng2023u}} & UIEB~\cite{li2019underwater} & 18.29 & 0.7650 & 0.3186 \\
 & LSUI~\cite{peng2023u} & 19.31 & 0.8017 & \textbf{0.2530} \\
 & EUVP~\cite{islam2020fast} & \textbf{20.65} & \textbf{0.8046} & 0.2837 \\
 & UWImgNetSD & 16.89 & 0.6767 & 0.3438 \\
 & UWNature (O) & 18.34 & 0.7665 & 0.3243 \\
 & UWNature (RV) & 18.73 & 0.7700 & 0.2965 \\
\midrule
\multirow{6}{*}{DMunderwater~\cite{tang2023underwater}} & UIEB~\cite{li2019underwater} & 15.44 & 0.7592 & 0.3210 \\
 & LSUI~\cite{peng2023u} & \textbf{22.93} & \textbf{0.8447} & \textbf{0.1800} \\
 & EUVP~\cite{islam2020fast} & 21.82 & 0.8344 & 0.1901 \\
 & UWImgNetSD & 17.87 & 0.7464 & 0.3242 \\
 & UWNature (O) & 19.36 & 0.7953 & 0.2708 \\
 & UWNature (RV) & 16.99 & 0.7412 & 0.3169 \\
\midrule
\multirow{6}{*}{WF-Diff~\cite{zhao2024wavelet}} & UIEB~\cite{li2019underwater} & 19.63 & 0.8141 & 0.2491 \\
 & LSUI~\cite{peng2023u} & \textbf{23.59} & 0.8587 & \textbf{0.1783} \\
 & EUVP~\cite{islam2020fast} & 23.37 & \textbf{0.8604} & 0.1932 \\
 & UWImgNetSD & 18.04 & 0.7481 & 0.3196 \\
 & UWNature (O) & 19.16 & 0.7932 & 0.2758 \\
 & UWNature (RV) & 18.48 & 0.7663 & 0.2935 \\
\bottomrule
\end{tabular}%
}
\end{table}

\subsubsection{UCIQE components}
UCIQE combines chroma variation $\sigma_c$, luminance contrast $C_L$, and mean saturation $\mu_s$ as
\begin{equation}
\mathrm{UCIQE}=0.4680\sigma_c+0.2745C_L+0.2576\mu_s.
\label{eq:sup-uciqe}
\end{equation}
\begin{table*}[!t]
\centering
\caption{UCIQE and its components for the 6 restoration architectures on U45, RUIE, and LSUI\&EUVP.}
\label{tab:sup-uciqe}
\setlength{\tabcolsep}{1.2pt}
\renewcommand{\arraystretch}{0.90}
\scriptsize
\resizebox{0.95\textwidth}{!}{%
\begin{tabular}{ll*{4}{ccc}}
\toprule
\multirow[c]{3}{*}{Model} &
\multirow[c]{3}{*}{Training Set} &
\multicolumn{3}{c}{UCIQE $\uparrow$} & \multicolumn{3}{c}{$\sigma_c$} & \multicolumn{3}{c}{$C_L$} & \multicolumn{3}{c}{$\mu_s$} \\
\cmidrule(lr){3-5}\cmidrule(lr){6-8}\cmidrule(lr){9-11}\cmidrule(lr){12-14}
 & & \multirow[c]{2}{*}{U45} & \multirow[c]{2}{*}{RUIE} & LSUI & \multirow[c]{2}{*}{U45} & \multirow[c]{2}{*}{RUIE} & LSUI & \multirow[c]{2}{*}{U45} & \multirow[c]{2}{*}{RUIE} & LSUI & \multirow[c]{2}{*}{U45} & \multirow[c]{2}{*}{RUIE} & LSUI \\
& &  &  & \&EUVP &  &  & \&EUVP &  &  & \&EUVP &  &  & \&EUVP \\
\midrule
\multirow{5}{*}{UWCNN~\cite{li2020underwater}} & UIEB~\cite{li2019underwater} & 0.5507 & 0.5115 & 0.5732 & 0.2607 & 0.1844 & 0.2628 & 0.8026 & 0.7584 & 0.8596 & 0.8090 & 0.8425 & 0.8317 \\
 & LSUI~\cite{peng2023u} & 0.5390 & 0.5113 & 0.5773 & 0.2489 & 0.1944 & 0.2692 & 0.7843 & 0.7844 & 0.8677 & 0.8045 & 0.7957 & 0.8273 \\
 & EUVP~\cite{islam2020fast} & 0.5369 & 0.5177 & 0.5745 & 0.2666 & 0.2128 & 0.2769 & 0.7380 & 0.7355 & 0.8324 & 0.8135 & 0.8394 & 0.8404 \\
 & UWImgNetSD & \textbf{0.5612} & \textbf{0.5493} & \textbf{0.5979} & 0.2957 & 0.2555 & 0.3053 & 0.7641 & 0.7689 & 0.8557 & 0.8269 & 0.8488 & 0.8544 \\
 & UWNature (O) & 0.5485 & 0.5213 & 0.5798 & 0.2904 & 0.2192 & 0.2904 & 0.7236 & 0.7278 & 0.8141 & 0.8307 & 0.8498 & 0.8556 \\
\midrule
\multirow{5}{*}{UIEC$^2$-Net~\cite{wang2021uiec}} & UIEB~\cite{li2019underwater} & \textbf{0.6172} & 0.5726 & 0.6121 & 0.3346 & 0.2661 & 0.3064 & 0.9122 & 0.8841 & 0.9342 & 0.8161 & 0.7972 & 0.8238 \\
 & LSUI~\cite{peng2023u} & 0.5785 & 0.5554 & 0.5835 & 0.2730 & 0.2383 & 0.2636 & 0.8480 & 0.8501 & 0.8815 & 0.8460 & 0.8172 & 0.8470 \\
 & EUVP~\cite{islam2020fast} & 0.5302 & 0.4960 & 0.5745 & 0.2599 & 0.2031 & 0.2661 & 0.7123 & 0.6869 & 0.8474 & 0.8268 & 0.8245 & 0.8438 \\
 & UWImgNetSD & 0.5989 & \textbf{0.5796} & \textbf{0.6310} & 0.2993 & 0.2551 & 0.3252 & 0.8868 & 0.8795 & 0.9450 & 0.8362 & 0.8494 & 0.8516 \\
 & UWNature (O) & 0.5810 & 0.5480 & 0.6035 & 0.2721 & 0.2175 & 0.2919 & 0.8431 & 0.8429 & 0.8891 & 0.8628 & 0.8340 & 0.8648 \\
\midrule
\multirow{5}{*}{WaterNet~\cite{li2019underwater}} & UIEB~\cite{li2019underwater} & 0.5480 & 0.5154 & 0.5756 & 0.2564 & 0.1847 & 0.2618 & 0.8011 & 0.7817 & 0.8777 & 0.8077 & 0.8324 & 0.8237 \\
 & LSUI~\cite{peng2023u} & 0.5315 & 0.5116 & 0.5689 & 0.2524 & 0.1999 & 0.2640 & 0.7344 & 0.7375 & 0.8275 & 0.8223 & 0.8370 & 0.8470 \\
 & EUVP~\cite{islam2020fast} & 0.5353 & 0.5181 & 0.5722 & 0.2672 & 0.2097 & 0.2775 & 0.7332 & 0.7460 & 0.8227 & 0.8111 & 0.8353 & 0.8405 \\
 & UWImgNetSD & 0.5549 & 0.5412 & 0.5958 & 0.2778 & 0.2405 & 0.2868 & 0.7585 & 0.7413 & 0.8771 & 0.8411 & 0.8741 & 0.8572 \\
 & UWNature (O) & 0.5730 & 0.5416 & 0.5897 & 0.3105 & 0.2397 & 0.2988 & 0.7690 & 0.7666 & 0.8332 & 0.8408 & 0.8503 & 0.8585 \\
\midrule
\multirow{5}{*}{U-shape~\cite{peng2023u}} & UIEB~\cite{li2019underwater} & \textbf{0.5964} & 0.5529 & 0.6012 & 0.3257 & 0.2602 & 0.3070 & 0.8669 & 0.8241 & 0.9010 & 0.7999 & 0.7956 & 0.8160 \\
 & LSUI~\cite{peng2023u} & 0.5654 & 0.5489 & 0.5861 & 0.2804 & 0.2449 & 0.2827 & 0.8287 & 0.8583 & 0.8916 & 0.8026 & 0.7715 & 0.8117 \\
 & EUVP~\cite{islam2020fast} & 0.5136 & 0.5159 & 0.5638 & 0.2421 & 0.2184 & 0.2711 & 0.7042 & 0.7291 & 0.8086 & 0.8037 & 0.8289 & 0.8345 \\
 & UWImgNetSD & 0.5833 & \textbf{0.5641} & \textbf{0.6250} & 0.3129 & 0.2566 & 0.3281 & 0.8060 & 0.8025 & 0.9083 & 0.8370 & 0.8685 & 0.8620 \\
 & UWNature (O) & 0.5734 & 0.5517 & 0.6017 & 0.3225 & 0.2707 & 0.3385 & 0.7638 & 0.7576 & 0.8221 & 0.8259 & 0.8424 & 0.8446 \\
\midrule
\multirow{5}{*}{DMunderwater~\cite{tang2023underwater}} & UIEB~\cite{li2019underwater} & 0.5854 & 0.5430 & 0.5802 & 0.3037 & 0.2372 & 0.2831 & 0.8953 & 0.8645 & 0.9066 & 0.7668 & 0.7556 & 0.7718 \\
 & LSUI~\cite{peng2023u} & 0.5827 & 0.5700 & 0.5853 & 0.2862 & 0.2592 & 0.2913 & 0.8546 & 0.8759 & 0.8553 & 0.8313 & 0.8085 & 0.8316 \\
 & EUVP~\cite{islam2020fast} & 0.5124 & 0.5020 & 0.5650 & 0.2385 & 0.2034 & 0.2682 & 0.7069 & 0.7174 & 0.8220 & 0.8025 & 0.8148 & 0.8301 \\
 & UWImgNetSD & \textbf{0.5957} & \textbf{0.5710} & \textbf{0.6131} & 0.3271 & 0.2755 & 0.3371 & 0.8355 & 0.8193 & 0.8722 & 0.8280 & 0.8431 & 0.8379 \\
 & UWNature (O) & 0.5667 & 0.5270 & 0.5887 & 0.3031 & 0.2247 & 0.3100 & 0.7706 & 0.7496 & 0.8230 & 0.8279 & 0.8386 & 0.8451 \\
\midrule
\multirow{5}{*}{WF-Diff~\cite{zhao2024wavelet}} & UIEB~\cite{li2019underwater} & \textbf{0.6026} & 0.5552 & 0.6038 & 0.3330 & 0.2605 & 0.3106 & 0.8438 & 0.7783 & 0.8708 & 0.8349 & 0.8527 & 0.8518 \\
 & LSUI~\cite{peng2023u} & 0.5786 & \textbf{0.5687} & 0.5869 & 0.2909 & 0.2677 & 0.2926 & 0.8274 & 0.8535 & 0.8534 & 0.8360 & 0.8118 & 0.8371 \\
 & EUVP~\cite{islam2020fast} & 0.5332 & 0.5083 & 0.5725 & 0.2616 & 0.1909 & 0.2765 & 0.7295 & 0.7432 & 0.8208 & 0.8171 & 0.8342 & 0.8455 \\
 & UWImgNetSD & 0.5931 & 0.5662 & \textbf{0.6120} & 0.3400 & 0.2788 & 0.3413 & 0.7952 & 0.7855 & 0.8514 & 0.8374 & 0.8543 & 0.8486 \\
 & UWNature (O) & 0.5761 & 0.5369 & 0.5998 & 0.3122 & 0.2432 & 0.3286 & 0.7811 & 0.7459 & 0.8229 & 0.8366 & 0.8477 & 0.8545 \\
\bottomrule
\end{tabular}%
}
\end{table*}

Table~\ref{tab:sup-uciqe} reveals that the synthetic training sources achieve the largest $\sigma_c$ in 15 of the 18 architecture--test set conditions and the largest $\mu_s$ in 17, whereas UIEB leads $C_L$ in 11 conditions.
The strong UCIQE results of synthetic supervision thus arise more consistently from the distribution and saturation of restored colors than from maximizing luminance contrast.
The frequent contrast advantage of UIEB is consistent with the visually enhanced references used in its training data, which can encourage a different balance of restoration attributes.
For DMunderwater on RUIE, UWImgNetSD and LSUI obtain similar UCIQE scores (0.5710 and 0.5700), but the former produces higher chroma variation (0.2755 versus 0.2592) and saturation (0.8431 versus 0.8085), while the latter produces higher luminance contrast (0.8759 versus 0.8193).
Nearly identical total scores can therefore reflect different image characteristics.
Across all 6 architectures on LSUI\&EUVP, UWImgNetSD obtains the highest UCIQE among the original training sources, consistent with its strong chroma and saturation responses.
These component differences help explain the restoration appearances in Figs.~5--7, although stronger chroma or saturation alone does not determine visual naturalness.

\subsubsection{UIQM components}
UIQM measures a related but different combination of colorfulness (UICM), sharpness (UISM), and contrast (UIConM).
For the evaluated implementation, the 3 components are combined as
\begin{equation}
\mathrm{UIQM}=0.0282\,\mathrm{UICM}+0.2953\,\mathrm{UISM}+3.5753\,\mathrm{UIConM}.
\label{eq:sup-uiqm}
\end{equation}
\begin{table*}[!t]
\centering
\caption{UIQM and its components for the 6 restoration architectures on U45, RUIE, and LSUI\&EUVP.}
\label{tab:sup-uiqm}
\setlength{\tabcolsep}{1.2pt}
\renewcommand{\arraystretch}{0.90}
\scriptsize
\resizebox{0.95\textwidth}{!}{%
\begin{tabular}{ll*{4}{ccc}}
\toprule
\multirow[c]{3}{*}{Model} &
\multirow[c]{3}{*}{Training Set} &
\multicolumn{3}{c}{UIQM $\uparrow$} & \multicolumn{3}{c}{UICM} & \multicolumn{3}{c}{UISM} & \multicolumn{3}{c}{UIConM} \\
\cmidrule(lr){3-5}\cmidrule(lr){6-8}\cmidrule(lr){9-11}\cmidrule(lr){12-14}
 & & \multirow[c]{2}{*}{U45} & \multirow[c]{2}{*}{RUIE} & LSUI & \multirow[c]{2}{*}{U45} & \multirow[c]{2}{*}{RUIE} & LSUI & \multirow[c]{2}{*}{U45} & \multirow[c]{2}{*}{RUIE} & LSUI & \multirow[c]{2}{*}{U45} & \multirow[c]{2}{*}{RUIE} & LSUI \\
& &  &  & \&EUVP &  &  & \&EUVP &  &  & \&EUVP &  &  & \&EUVP \\
\midrule
\multirow{5}{*}{UWCNN~\cite{li2020underwater}} & UIEB~\cite{li2019underwater} & \textbf{0.688} & 0.347 & 0.795 & 6.730 & 3.354 & 5.893 & 4.765 & 3.875 & 5.029 & -0.2542 & -0.2495 & -0.2394 \\
 & LSUI~\cite{peng2023u} & 0.646 & \textbf{0.746} & \textbf{0.980} & 6.479 & 10.186 & 10.196 & 4.452 & 4.445 & 5.175 & -0.2383 & -0.2389 & -0.2339 \\
 & EUVP~\cite{islam2020fast} & 0.454 & 0.334 & 0.746 & 4.260 & 4.705 & 7.957 & 3.955 & 3.498 & 4.721 & -0.2334 & -0.2326 & -0.2441 \\
 & UWImgNetSD & 0.639 & 0.547 & 0.913 & 5.765 & 9.098 & 9.549 & 4.416 & 3.884 & 4.871 & -0.2315 & -0.2396 & -0.2222 \\
 & UWNature (O) & 0.613 & 0.608 & 0.882 & 5.505 & 9.990 & 10.130 & 4.449 & 4.095 & 4.961 & -0.2395 & -0.2469 & -0.2429 \\
\midrule
\multirow{5}{*}{UIEC$^2$-Net~\cite{wang2021uiec}} & UIEB~\cite{li2019underwater} & 0.797 & 0.823 & 0.937 & 9.491 & 12.582 & 10.963 & 4.810 & 4.537 & 5.111 & -0.2492 & -0.2438 & -0.2465 \\
 & LSUI~\cite{peng2023u} & 0.655 & 0.773 & 0.866 & 10.925 & 14.071 & 11.865 & 4.152 & 4.225 & 4.798 & -0.2460 & -0.2437 & -0.2475 \\
 & EUVP~\cite{islam2020fast} & 0.471 & 0.289 & 0.761 & 4.576 & 1.344 & 7.659 & 4.096 & 3.767 & 4.821 & -0.2428 & -0.2410 & -0.2457 \\
 & UWImgNetSD & \textbf{1.101} & \textbf{0.950} & \textbf{1.286} & 12.632 & 11.289 & 12.643 & 5.230 & 4.891 & 5.559 & -0.2236 & -0.2274 & -0.1991 \\
 & UWNature (O) & 0.737 & 0.799 & 0.853 & 10.677 & 12.749 & 12.076 & 4.450 & 4.488 & 4.706 & -0.2456 & -0.2478 & -0.2454 \\
\midrule
\multirow{5}{*}{WaterNet~\cite{li2019underwater}} & UIEB~\cite{li2019underwater} & 0.671 & 0.398 & 0.788 & 6.336 & 3.855 & 5.576 & 4.744 & 3.955 & 5.082 & -0.2541 & -0.2457 & -0.2432 \\
 & LSUI~\cite{peng2023u} & 0.584 & 0.642 & \textbf{0.990} & 5.816 & 11.269 & 10.895 & 4.361 & 4.017 & 5.085 & -0.2426 & -0.2411 & -0.2289 \\
 & EUVP~\cite{islam2020fast} & 0.462 & 0.344 & 0.758 & 4.182 & 5.016 & 7.977 & 3.995 & 3.477 & 4.753 & -0.2337 & -0.2306 & -0.2435 \\
 & UWImgNetSD & 0.524 & 0.412 & 0.918 & 6.385 & 8.999 & 8.471 & 4.084 & 3.484 & 4.876 & -0.2411 & -0.2436 & -0.2127 \\
 & UWNature (O) & 0.635 & 0.630 & 0.869 & 5.554 & 10.547 & 10.750 & 4.576 & 4.112 & 4.885 & -0.2443 & -0.2467 & -0.2453 \\
\midrule
\multirow{5}{*}{U-shape~\cite{peng2023u}} & UIEB~\cite{li2019underwater} & 0.614 & 0.569 & 0.825 & 5.548 & 4.849 & 6.532 & 4.601 & 4.454 & 5.089 & -0.2520 & -0.2469 & -0.2410 \\
 & LSUI~\cite{peng2023u} & 0.668 & \textbf{0.883} & 0.923 & 9.560 & 12.817 & 9.477 & 4.309 & 4.651 & 5.124 & -0.2446 & -0.2383 & -0.2399 \\
 & EUVP~\cite{islam2020fast} & 0.390 & 0.364 & 0.732 & 3.462 & 6.131 & 8.512 & 3.751 & 3.455 & 4.625 & -0.2281 & -0.2319 & -0.2443 \\
 & UWImgNetSD & \textbf{0.676} & 0.815 & \textbf{1.154} & 5.707 & 11.963 & 10.956 & 4.577 & 4.473 & 5.188 & -0.2340 & -0.2360 & -0.1922 \\
 & UWNature (O) & 0.627 & 0.717 & 0.850 & 5.632 & 9.022 & 10.244 & 4.572 & 4.651 & 4.881 & -0.2468 & -0.2549 & -0.2462 \\
\midrule
\multirow{5}{*}{DMunderwater~\cite{tang2023underwater}} & UIEB~\cite{li2019underwater} & 0.625 & 0.557 & 0.714 & 3.268 & 3.123 & 3.211 & 4.862 & 4.478 & 5.181 & -0.2526 & -0.2386 & -0.2536 \\
 & LSUI~\cite{peng2023u} & 0.635 & \textbf{0.774} & 0.855 & 9.377 & 13.202 & 9.217 & 4.148 & 4.279 & 4.944 & -0.2388 & -0.2409 & -0.2421 \\
 & EUVP~\cite{islam2020fast} & 0.514 & 0.375 & 0.753 & 3.995 & 5.511 & 8.936 & 3.999 & 3.251 & 4.665 & -0.2181 & -0.2072 & -0.2451 \\
 & UWImgNetSD & \textbf{0.790} & 0.768 & \textbf{0.972} & 7.968 & 9.235 & 10.400 & 4.578 & 4.371 & 5.020 & -0.2199 & -0.2190 & -0.2248 \\
 & UWNature (O) & 0.641 & 0.702 & 0.833 & 6.068 & 10.122 & 10.038 & 4.521 & 4.378 & 4.802 & -0.2420 & -0.2450 & -0.2430 \\
\midrule
\multirow{5}{*}{WF-Diff~\cite{zhao2024wavelet}} & UIEB~\cite{li2019underwater} & 0.672 & 0.486 & 0.821 & 8.218 & 9.743 & 9.544 & 4.471 & 3.637 & 4.660 & -0.2462 & -0.2412 & -0.2305 \\
 & LSUI~\cite{peng2023u} & 0.667 & \textbf{0.807} & 0.860 & 10.222 & 14.040 & 9.747 & 4.183 & 4.319 & 4.838 & -0.2395 & -0.2417 & -0.2359 \\
 & EUVP~\cite{islam2020fast} & 0.484 & 0.304 & 0.714 & 5.057 & 5.281 & 8.912 & 3.820 & 3.092 & 4.489 & -0.2201 & -0.2120 & -0.2413 \\
 & UWImgNetSD & 0.771 & 0.679 & \textbf{0.938} & 7.404 & 9.144 & 10.682 & 4.612 & 4.158 & 4.848 & -0.2236 & -0.2257 & -0.2224 \\
 & UWNature (O) & 0.677 & 0.614 & 0.823 & 6.910 & 9.727 & 10.000 & 4.508 & 4.046 & 4.648 & -0.2373 & -0.2390 & -0.2325 \\
\bottomrule
\end{tabular}%
}
\end{table*}

The different component preferences become especially clear on RUIE, where LSUI gives the highest UIQM for five of the 6 architectures among the original training sources, despite the strong learned perceptual scores of the synthetic datasets in Section V-C.
Its UICM is the highest in 11 of the 18 architecture--test set conditions, indicating that its UIQM advantage is often associated with the metric's colorfulness response rather than with uniformly higher contrast.
Averaged equally over the 6 networks on RUIE, LSUI exceeds UWImgNetSD in UIQM by 0.076; the weighted UICM and UISM differences contribute approximately $+0.075$ and $+0.033$, while the UIConM difference contributes $-0.032$.
The situation reverses on LSUI\&EUVP, where UWImgNetSD exceeds LSUI by 0.118 in mean UIQM and the UIConM term accounts for approximately 0.092 of that difference.
The contrast component is negative under this UIQM implementation, so its contribution is interpreted through the weighted score difference rather than as a direct measure of physical contrast.
Taken together, the UCIQE and UIQM decompositions show why handcrafted underwater metrics can favor different training sources on the same images: they reward different combinations of color spread, saturation, sharpness, and contrast.
The complementary learned perceptual scores and MOS in the main manuscript provide a broader assessment of the visual quality produced by these supervision sources.

\subsubsection{Effect of the appearance targets}
\begin{figure*}[!t]
\centering
\includegraphics[width=0.95\textwidth]{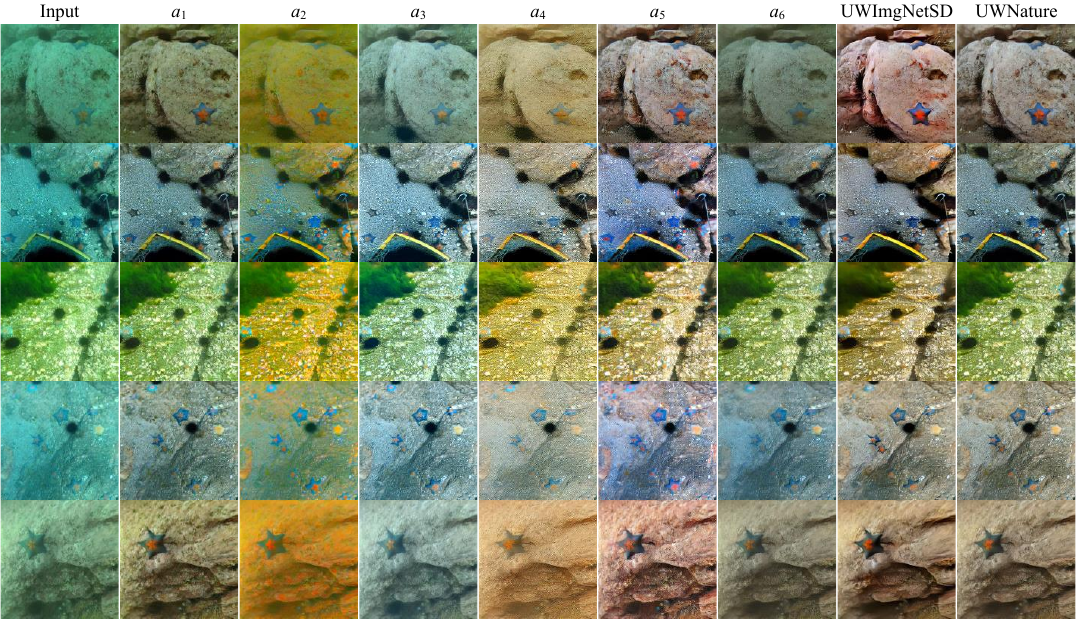}
\caption{DMunderwater outputs on RUIE after training with individual appearance targets ($a_1$--$a_6$) or combined-target datasets.
The reddish--violet cast from $a_5$ and yellow--orange cast from $a_2$ contrast with the more balanced outputs from mixed-target supervision.}
\label{fig:sup-appearance-ablation}
\end{figure*}

The framework uses 6 visually distinctive underwater appearances to guide independent I2IT models, as described in Section III-B.
An additional experiment trains DMunderwater using synthetic pairs from one target domain at a time, with the same source image subset and training procedure.
The indices $a_1$--$a_6$ denote the 6 representative appearance targets shown in Fig.~1.
Table~\ref{tab:sup-anchor-ablation} reports UCIQE and its components on RUIE for the 6 individual appearance targets and the two datasets assembled from all 6 targets.
The single-target results vary substantially, from 0.4895 for $a_3$ to 0.5465 for $a_5$.
The qualitative comparison in Fig.~\ref{fig:sup-appearance-ablation} explains why this ranking cannot be interpreted as a ranking of color restoration accuracy.
The $a_5$ model, trained on a green-toned appearance target, introduces conspicuous reddish and violet color casts on several rocks and seabed regions, even though it achieves the highest UCIQE among the 6 single-target models.
Relative to UWNature, $a_5$ has greater chroma variation (0.2602 versus 0.2247) and luminance contrast (0.7641 versus 0.7496), which raise its UCIQE without necessarily producing more plausible colors.
The $a_2$ model illustrates the same limitation from another direction, imposing a strong yellow-orange appearance across scenes that differ in their original underwater color casts.
These examples suggest that a model trained on one appearance target can learn a characteristic color correction that transfers poorly to other underwater appearances.
By comparison, the models trained on the 6 target datasets produce more balanced colors across the illustrated scenes, without applying the same pronounced cast to every input.
UWImgNetSD achieves 0.5710 UCIQE, above all 6 single-target models, while UWNature scores 0.5270, below $a_5$ despite its more restrained coloration in the qualitative examples.
The comparison therefore favors appearance diversity as a way to reduce dependence on a single learned color style, rather than as a guarantee of the highest score under every handcrafted metric.
The difference between UWNature and UWImgNetSD also indicates that the clear reference source influences the resulting restoration, even when the same 6 I2IT target models are used.
Together with the multi-metric and perceptual evaluation in Section V, this supports the data construction strategy in Section III: representative target appearances provide varied synthetic inputs, while the original clear images remain stable restoration references.

\begin{table}[h]
\caption{UCIQE and its components for DMunderwater on RUIE using individual I2IT appearance targets and the datasets constructed from all 6 targets.}
\label{tab:sup-anchor-ablation}
\centering\scriptsize\setlength{\tabcolsep}{3pt}
\begin{tabular}{lrrrr}\toprule
Target & UCIQE $\uparrow$ & $\sigma_c$ & $C_L$ & $\mu_s$ \\
\midrule
$a_1$ & 0.5159 & 0.1927 & 0.7429 & 0.8610 \\
$a_2$ & 0.5170 & 0.2495 & 0.6671 & 0.8429 \\
$a_3$ & 0.4895 & 0.1801 & 0.7518 & 0.7718 \\
$a_4$ & 0.4914 & 0.1955 & 0.7121 & 0.7938 \\
$a_5$ & 0.5465 & 0.2602 & 0.7641 & 0.8348 \\
$a_6$ & 0.5260 & 0.1890 & 0.6861 & 0.8547 \\
\midrule
UWImgNetSD & 0.5710 & 0.2755 & 0.8193 & 0.8431 \\
UWNature & 0.5270 & 0.2247 & 0.7496 & 0.8386 \\
\bottomrule\end{tabular}
\end{table}

\subsection{Human Perceptual Evaluation}
Table~\ref{tab:sup-mos-groups} reports MOS for each combination of restoration architecture and test set, complementing the overall scores in Table V.
Each source is evaluated on 8 U45 scenes, 5 RUIE scenes, and 14 LSUI\&EUVP scenes per architecture, with two ratings per stimulus.
The joint grouping helps distinguish broadly preferred supervision from architecture-specific responses to reference appearance.
\begin{table}[h]
\caption{MOS grouped jointly by restoration architecture and test set, with five training sources compared on the same scenes.}
\label{tab:sup-mos-groups}
\centering\scriptsize\setlength{\tabcolsep}{2.0pt}
\begin{tabular}{llrrrrr}\toprule
Model & Test set & UIEB & LSUI & EUVP & Nature & ImgNetSD \\
\midrule
UWCNN & U45 & 3.125 & 3.312 & 3.062 & 3.438 & 3.562 \\
 & RUIE & 3.400 & 3.800 & 4.200 & 4.200 & 3.900 \\
 & LSUI\&EUVP & 3.464 & 3.607 & 3.321 & 3.679 & 3.786 \\
\midrule
U-shape & U45 & 2.750 & 2.312 & 2.125 & 3.062 & 3.312 \\
 & RUIE & 3.800 & 3.800 & 3.900 & 4.000 & 4.400 \\
 & LSUI\&EUVP & 2.964 & 3.250 & 3.000 & 3.750 & 3.179 \\
\midrule
WF-Diff & U45 & 3.750 & 3.875 & 2.875 & 3.562 & 3.812 \\
 & RUIE & 3.400 & 4.100 & 3.900 & 4.200 & 4.800 \\
 & LSUI\&EUVP & 3.500 & 3.607 & 3.786 & 3.643 & 4.143 \\
\bottomrule\end{tabular}\end{table}

The grouped MOS results show that the overall advantage of the two synthetic sources is not confined to one restoration architecture, but their relative preference varies with the model and test scene.
On LSUI\&EUVP, U-shape trained on UWNature scores 3.750 compared with 3.179 for UWImgNetSD, whereas WF-Diff scores 3.643 and 4.143, respectively.
Thus, the reference appearance that benefits one architecture need not be preferred by another, even under the same test distribution.
Among the 70 repeated rating pairs, 57.1\% receive identical scores and 91.4\% differ by at most one point, with a mean absolute difference of 0.514.
The leave-one-participant-out analysis preserves the overall ordering of UWImgNetSD first and UWNature second in all ten omissions, although their matched mean difference remains small relative to participant variation.
These observations complement Table V by showing both the consistency of the overall perceptual trend and the model-dependent preferences hidden by a single MOS average.

\subsection{Further Analysis of Restoration Behavior}
\subsubsection{Depth modulation}
Section V-F compares UWNature (O) with its depth modulation variant (RV) through 20,000 paired bootstrap resamples for each evaluation condition.
Table~\ref{tab:sup-or-family} groups the 144 comparisons by architecture and test set.
For each architecture, U45 and RUIE contribute seven no-reference metrics each, and LSUI\&EUVP contributes seven no-reference and 3 full-reference metrics.
The results distinguish changes in the spatial plausibility of synthetic inputs from their usefulness for learning the restoration mapping.
\begin{table}[h]
\caption{Outcomes of the paired O/RV comparison for each architecture and test set.}
\label{tab:sup-or-family}
\centering\scriptsize\setlength{\tabcolsep}{3.0pt}
\begin{tabular}{llrrrr}\toprule
Model & Test set & RV & Unclear & O & Total \\
\midrule
UWCNN & U45 & 0 & 1 & 6 & 7 \\
 & RUIE & 1 & 0 & 6 & 7 \\
 & LSUI\&EUVP & 2 & 0 & 8 & 10 \\
\midrule
UIEC$^2$-Net & U45 & 1 & 4 & 2 & 7 \\
 & RUIE & 0 & 1 & 6 & 7 \\
 & LSUI\&EUVP & 5 & 1 & 4 & 10 \\
\midrule
WaterNet & U45 & 1 & 2 & 4 & 7 \\
 & RUIE & 2 & 0 & 5 & 7 \\
 & LSUI\&EUVP & 3 & 1 & 6 & 10 \\
\midrule
U-shape & U45 & 5 & 0 & 2 & 7 \\
 & RUIE & 5 & 0 & 2 & 7 \\
 & LSUI\&EUVP & 7 & 0 & 3 & 10 \\
\midrule
DMunderwater & U45 & 2 & 4 & 1 & 7 \\
 & RUIE & 5 & 0 & 2 & 7 \\
 & LSUI\&EUVP & 3 & 2 & 5 & 10 \\
\midrule
WF-Diff & U45 & 3 & 2 & 2 & 7 \\
 & RUIE & 4 & 0 & 3 & 7 \\
 & LSUI\&EUVP & 2 & 0 & 8 & 10 \\
\midrule
Total & & 51 & 18 & 75 & 144 \\
\bottomrule\end{tabular}\end{table}

\begin{figure*}[!t]
\centering
\includegraphics[width=0.9\textwidth]{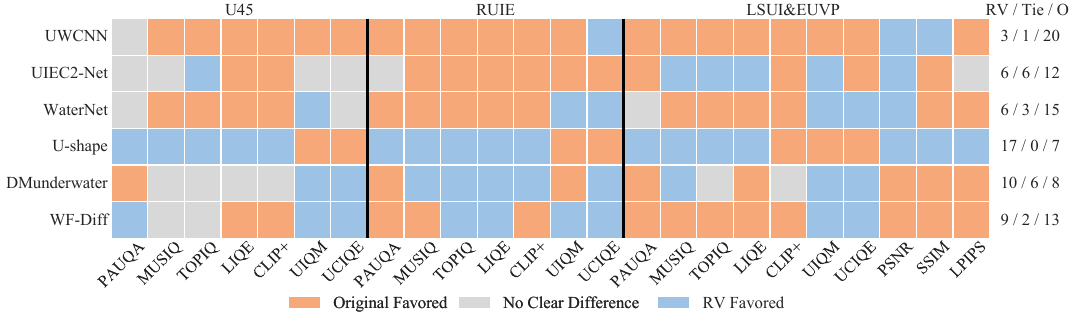}
\caption{Paired-bootstrap comparison of UWNature (O) and depth-modulated UWNature (RV) across 6 architectures and 144 evaluation conditions.
Within each test-set block, R indicates RV favored, O indicates original favored, and a dot indicates no clear difference under the 95\% confidence interval.
The distinct model- and test-set patterns show the nonuniform effects of depth modulation.}
\label{fig:sup-or-matrix}
\end{figure*}

The outcomes are strongly architecture-dependent: RV is favored in 17 of 24 conditions for U-shape, spanning all 3 test sets, but in only 3 of 24 conditions for UWCNN.
Fig.~\ref{fig:sup-or-matrix} reveals the structure behind the aggregate counts: U-shape has RV-favored cells in every test-set block, whereas O-favored cells dominate the UWCNN row.
The pattern is not fixed by architecture alone, since WF-Diff favors RV in 4 of 7 RUIE comparisons but in only 2 of 10 LSUI\&EUVP comparisons.
The distinct blocks in the matrix indicate that a depth-dependent input distribution interacts with both the restoration model and the evaluation domain, rather than producing a uniform gain in color, perceptual quality, or reference agreement.
U-shape may be better able to use the added spatial variation, while the original input distribution can remain more favorable for other architectures; the present comparisons establish the performance difference without isolating an architectural mechanism.
This complements the spatial controls above and the central finding of Section V-F: a more plausible spatial pattern of degradation does not necessarily lead to more effective supervision for every learned restoration mapping.

\subsubsection{Degradation and restoration reciprocity}
Table~\ref{tab:sup-reciprocity} compares the local estimate of the inverse synthetic degradation direction with global, random, and shuffled estimates.
For U-shape trained on UWNature (O), the local estimate gives a higher mean cosine similarity on all 3 test sets.
Its macro mean of 0.910 exceeds the global estimate of 0.861 and the randomized controls, indicating that matching similar synthetic degraded appearances adds information beyond a common average correction direction.
The macro mean assigns equal weight to each test set.

\begin{table}[h]
\caption{Mean cosine similarity between estimated inverse degradation and restoration directions for U-shape trained on UWNature (O).}
\label{tab:sup-reciprocity}
\centering\scriptsize\setlength{\tabcolsep}{2.6pt}
\begin{tabular}{lrrrr}\toprule
Test set & Local $K=5$ & Global & Random & Shuffled \\
\midrule
U45 & 0.870 & 0.830 & 0.786 & 0.777 \\
RUIE & 0.958 & 0.897 & 0.849 & 0.906 \\
LSUI\&EUVP & 0.902 & 0.855 & 0.814 & 0.792 \\
\midrule
Macro mean & 0.910 & 0.861 & 0.816 & 0.825 \\
\bottomrule\end{tabular}\end{table}

The descriptor ablation in Table~\ref{tab:sup-attribute} identifies which appearance changes contribute most to that agreement, with the restoration model held fixed.
On LSUI\&EUVP, color-only matching reaches a cosine similarity of 0.963, while removing the two CIELab chromatic coordinates reduces it to 0.669, compared with 0.902 for all five attributes.
The learned restoration is therefore particularly consistent with the inverse synthetic color changes, while luminance, contrast, and detail contribute less uniformly to the directional measure.
This complements the metric decomposition above: successful synthetic supervision involves a coherent correction of several visual attributes, but the strength of that correspondence varies with the attribute and the restoration architecture.

\begin{table}[h]
\caption{Mean directional cosine under different selections of appearance attributes for the same U-shape model.}
\label{tab:sup-attribute}
\centering\scriptsize\setlength{\tabcolsep}{4pt}
\begin{tabular}{lrrr}\toprule
Descriptor subset & U45 & RUIE & LSUI\&EUVP \\
\midrule
All five & 0.870 & 0.958 & 0.902 \\
Color only ($a^*,b^*$) & 0.885 & 0.981 & 0.963 \\
Without $a^*,b^*$ & 0.873 & 0.937 & 0.669 \\
Without $L^*$ & 0.893 & 0.969 & 0.935 \\
\bottomrule\end{tabular}\end{table}

The median directional similarity remains similar when the number of nearest synthetic samples increases from $K=5$ to $K=10$ or $K=20$.
Together, these controls show that the synthetic pairs provide an informative direction for the corrections learned on real underwater images, rather than merely reproducing a single global appearance shift.

%% file: ref.bib
@InProceedings{fu2022uncertainty,
    author="Fu, Zhenqi
        and Wang, Wu
        and Huang, Yue
        and Ding, Xinghao
        and Ma, Kai-Kuang",
    title="Uncertainty Inspired Underwater Image Enhancement",
    booktitle="Computer Vision -- ECCV 2022",
    year="2022",
    pages="465--482",
    isbn="978-3-031-19797-0"
}

@ARTICLE{li2018watergan,
  author={Li, Jie and Skinner, Katherine A. and Eustice, Ryan M. and Johnson-Roberson, Matthew},
  journal={IEEE Robotics and Automation Letters}, 
  title={{WaterGAN}: Unsupervised Generative Network to Enable Real-Time Color Correction of Monocular Underwater Images}, 
  year={2018},
  volume={3},
  number={1},
  pages={387-394},
  doi={10.1109/LRA.2017.2730363},
  month={Jan.}
}

@INPROCEEDINGS{akkaynak2019seathru,
  author={Akkaynak, Derya and Treibitz, Tali},
  booktitle={2019 IEEE/CVF Conference on Computer Vision and Pattern Recognition (CVPR)}, 
  title={Sea-Thru: A Method for Removing Water From Underwater Images}, 
  year={2019},
  volume={},
  number={},
  pages={1682-1691},
  doi={10.1109/CVPR.2019.00178}
}

@article{JIANG2022108324,
    title = {Two-step domain adaptation for underwater image enhancement},
    journal = {Pattern Recognition},
    volume = {122},
    pages = {108324},
    year = {2022},
    issn = {0031-3203},
    doi = {https://doi.org/10.1016/j.patcog.2021.108324},
    author = {Qun Jiang and Yunfeng Zhang and Fangxun Bao and Xiuyang Zhao and Caiming Zhang and Peide Liu}
}

@ARTICLE{10048777,
  author={Wang, Zhengyong and Shen, Liquan and Xu, Mai and Yu, Mei and Wang, Kun and Lin, Yufei},
  journal={IEEE Transactions on Image Processing}, 
  title={Domain Adaptation for Underwater Image Enhancement}, 
  year={2023},
  volume={32},
  number={},
  pages={1442-1457},
  doi={10.1109/TIP.2023.3244647}}

@ARTICLE{1284395,
  author={Zhou Wang and Bovik, A.C. and Sheikh, H.R. and Simoncelli, E.P.},
  journal={IEEE Transactions on Image Processing}, 
  title={Image quality assessment: from error visibility to structural similarity}, 
  year={2004},
  volume={13},
  number={4},
  pages={600-612},
  doi={10.1109/TIP.2003.819861}}

@INPROCEEDINGS{8578166,
  author={Zhang, Richard and Isola, Phillip and Efros, Alexei A. and Shechtman, Eli and Wang, Oliver},
  booktitle={2018 IEEE/CVF Conference on Computer Vision and Pattern Recognition}, 
  title={The Unreasonable Effectiveness of Deep Features as a Perceptual Metric}, 
  year={2018},
  volume={},
  number={},
  pages={586-595},
  doi={10.1109/CVPR.2018.00068}}

@ARTICLE{7300447,
  author={Yang, Miao and Sowmya, Arcot},
  journal={IEEE Transactions on Image Processing}, 
  title={An Underwater Color Image Quality Evaluation Metric}, 
  year={2015},
  volume={24},
  number={12},
  pages={6062-6071},
  doi={10.1109/TIP.2015.2491020}}

@ARTICLE{7305804,
  author={Panetta, Karen and Gao, Chen and Agaian, Sos},
  journal={IEEE Journal of Oceanic Engineering}, 
  title={Human-Visual-System-Inspired Underwater Image Quality Measures}, 
  year={2016},
  volume={41},
  number={3},
  pages={541-551},
  doi={10.1109/JOE.2015.2469915}}

@ARTICLE{11017762,
  author={Lin, Bosen and Dong, Junyu and Dong, Xinghui},
  journal={IEEE Transactions on Circuits and Systems for Video Technology}, 
  title={Perception-Aware Underwater Image Quality Assessment: Dataset, Perceptual Quality Scores, and Assessment Network}, 
  year={2025},
  volume={35},
  number={11},
  pages={11113-11128},
  doi={10.1109/TCSVT.2025.3574860}}

@INPROCEEDINGS{9710973,
  author={Ke, Junjie and Wang, Qifei and Wang, Yilin and Milanfar, Peyman and Yang, Feng},
  booktitle={2021 IEEE/CVF International Conference on Computer Vision (ICCV)}, 
  title={{MUSIQ}: Multi-scale Image Quality Transformer}, 
  year={2021},
  volume={},
  number={},
  pages={5128-5137},
  doi={10.1109/ICCV48922.2021.00510}}

@ARTICLE{10478301,
  author={Chen, Chaofeng and Mo, Jiadi and Hou, Jingwen and Wu, Haoning and Liao, Liang and Sun, Wenxiu and Yan, Qiong and Lin, Weisi},
  journal={IEEE Transactions on Image Processing}, 
  title={{TOPIQ}: A Top-Down Approach From Semantics to Distortions for Image Quality Assessment}, 
  year={2024},
  volume={33},
  number={},
  pages={2404-2418},
  doi={10.1109/TIP.2024.3378466}}

@INPROCEEDINGS{10204315,
  author={Zhang, Weixia and Zhai, Guangtao and Wei, Ying and Yang, Xiaokang and Ma, Kede},
  booktitle={2023 IEEE/CVF Conference on Computer Vision and Pattern Recognition (CVPR)}, 
  title={Blind Image Quality Assessment via Vision-Language Correspondence: A Multitask Learning Perspective}, 
  year={2023},
  volume={},
  number={},
  pages={14071-14081},
  doi={10.1109/CVPR52729.2023.01352}}

@inproceedings{10.1609/aaai.v37i2.25353,
    author = {Wang, Jianyi and Chan, Kelvin C.K. and Loy, Chen Change},
    title = {Exploring {CLIP} for assessing the look and feel of images},
    booktitle={Proceedings of the AAAI conference on artificial intelligence},
    year = {2023},
    isbn = {978-1-57735-880-0},
    doi = {10.1609/aaai.v37i2.25353},
    articleno = {284},
    pages={2555-2563},
    numpages = {9}
}

@inproceedings{dang2015raise,
  author = {Dang-Nguyen, Duc-Tien and Pasquini, Cecilia and Conotter, Valentina and Boato, Giulia},
  title = {{RAISE}: a raw images dataset for digital image forensics},
  year = {2015},
  booktitle = {Proceedings of the 6th ACM Multimedia Systems Conference},
  pages={219--224}
}

@INPROCEEDINGS{deng2009imagenet,
  author={Deng, Jia and Dong, Wei and Socher, Richard and Li, Li-Jia and Kai Li and Li Fei-Fei},
  booktitle={2009 IEEE Conference on Computer Vision and Pattern Recognition}, 
  title={{ImageNet}: A large-scale hierarchical image database}, 
  year={2009},
  volume={},
  number={},
  pages={248-255},
  doi={10.1109/CVPR.2009.5206848}
}

@INPROCEEDINGS{van2018inaturalist,
  author={Van Horn, Grant and Mac Aodha, Oisin and Song, Yang and Cui, Yin and Sun, Chen and Shepard, Alex and Adam, Hartwig and Perona, Pietro and Belongie, Serge},
  booktitle={2018 IEEE/CVF Conference on Computer Vision and Pattern Recognition}, 
  title={The {iNaturalist} Species Classification and Detection Dataset}, 
  year={2018},
  volume={},
  number={},
  pages={8769-8778},
  doi={10.1109/CVPR.2018.00914}}

@article{parmar2024one,
  title={One-step image translation with text-to-image models},
  author={Parmar, Gaurav and Park, Taesung and Narasimhan, Srinivasa and Zhu, Jun-Yan},
  journal={arXiv preprint arXiv:2403.12036},
  year={2024},
  month={Mar.}

}

@inproceedings{depth_anything_v2,
  author = {Yang, Lihe and Kang, Bingyi and Huang, Zilong and Zhao, Zhen and Xu, Xiaogang and Feng, Jiashi and Zhao, Hengshuang},
  title = {Depth anything {V2}},
  year = {2024},
  booktitle = {Proceedings of the 38th International Conference on Neural Information Processing Systems},
  volume = {37},
  pages = {21875-21911},
  doi = {10.52202/079017-0688}
}

@inproceedings{NEURIPS2025_336572db,
 author = {Wang, Ruicheng and Xu, Sicheng and Dong, Yue and Deng, Yu and Xiang, Jianfeng and Lv, Zelong and Sun, Guangzhong and Tong, Xin and Yang, Jiaolong},
 booktitle = {Advances in Neural Information Processing Systems},
 doi = {10.52202/085713-1207},
 pages = {35928--35959},
 title = {{MoGe-2}: Accurate Monocular Geometry with Metric Scale and Sharp Details},
 volume = {38, Main Conference},
 year = {2025}
}

@article{li2020underwater,
  title={Underwater scene prior inspired deep underwater image and video enhancement},
  author={Li, Chongyi and Anwar, Saeed and Porikli, Fatih},
  journal={Pattern Recognition},
  volume={98},
  pages={107038},
  year={2020},
  month={Sep.},
  publisher={Elsevier}
}

@article{wang2021uiec,
  author = {Wang, Yudong and Guo, Jichang and Gao, Huan and Yue, Huihui},
  title = {{UIEC$^{2}$-Net}: {CNN}-based underwater image enhancement using two color space},
  journal = {Signal Processing: Image Communication},
  year = {2021},
  volume = {96},
  pages = {116250},
  doi = {10.1016/j.image.2021.116250}
}

@ARTICLE{li2019underwater,
  author={Li, Chongyi and Guo, Chunle and Ren, Wenqi and Cong, Runmin and Hou, Junhui and Kwong, Sam and Tao, Dacheng},
  journal={IEEE Transactions on Image Processing}, 
  title={An Underwater Image Enhancement Benchmark Dataset and Beyond}, 
  year={2020},
  volume={29},
  number={},
  pages={4376-4389},
  month={Nov.},
  doi={10.1109/TIP.2019.2955241}
}

@article{islam2020fast,
  title={Fast underwater image enhancement for improved visual perception},
  author={Islam, Md Jahidul and Xia, Youya and Sattar, Junaed},
  journal={IEEE Robotics and Automation Letters},
  volume={5},
  number={2},
  pages={3227--3234},
  year={2020},
  month={Feb.},
  publisher={IEEE}
}

@article{peng2023u,
  title={U-shape transformer for underwater image enhancement},
  author={Peng, Lintao and Zhu, Chunli and Bian, Liheng},
  journal={IEEE Transactions on Image Processing},
  volume={32},
  pages={3066--3079},
  year={2023},
  month={May.},
  publisher={IEEE}
}

@article{li2019fusion,
  title={A fusion adversarial underwater image enhancement network with a public test dataset},
  author={Li, Hanyu and Li, Jingjing and Wang, Wei},
  journal={arXiv preprint arXiv:1906.06819},
  year={2019},
  month={Jun.}
}

@article{liu2020real,
  title={Real-world underwater enhancement: Challenges, benchmarks, and solutions under natural light},
  author={Liu, Risheng and Fan, Xin and Zhu, Ming and Hou, Minjun and Luo, Zhongxuan},
  journal={IEEE transactions on circuits and systems for video technology},
  volume={30},
  number={12},
  pages={4861--4875},
  year={2020},
  month={Dec.},
  publisher={IEEE}
}

@INPROCEEDINGS{zhao2024wavelet,
  author={Zhao, Chen and Cai, Weiling and Dong, Chenyu and Hu, Chengwei},
  booktitle={2024 IEEE/CVF Conference on Computer Vision and Pattern Recognition (CVPR)}, 
  title={Wavelet-based Fourier Information Interaction with Frequency Diffusion Adjustment for Underwater Image Restoration}, 
  year={2024},
  volume={},
  number={},
  pages={8281-8291},
  doi={10.1109/CVPR52733.2024.00791}
}

@inproceedings{tang2023underwater,
  title={Underwater image enhancement by transformer-based diffusion model with non-uniform sampling for skip strategy},
  author={Tang, Yi and Kawasaki, Hiroshi and Iwaguchi, Takafumi},
  booktitle={Proceedings of the 31st ACM International Conference on Multimedia},
  pages={5419--5427},
  year={2023},
  month={Oct.}
}

@INPROCEEDINGS{akkaynak2018formation,
  author={Akkaynak, Derya and Treibitz, Tali},
  booktitle={2018 IEEE/CVF Conference on Computer Vision and Pattern Recognition}, 
  title={A Revised Underwater Image Formation Model}, 
  year={2018},
  volume={},
  number={},
  pages={6723-6732},
  doi={10.1109/CVPR.2018.00703}}

@ARTICLE{li2025muse,
  author={Li, Songyang and Liu, Tingyu and Jiang, Qunyan and Li, Yuanqi and Guo, Jie and Jiao, Lei and Guo, Yanwen and Ni, Zhonghua},
  journal={IEEE Transactions on Geoscience and Remote Sensing}, 
  title={Realistic Simulation of Underwater Scene for Image Enhancement}, 
  year={2025},
  volume={63},
  number={},
  pages={1-14},
  month={Apr.},
  doi={10.1109/TGRS.2025.3561927}
  }

@ARTICLE{liu2025pseudoref,
  author={Liu, Yi and Jiang, Qiuping and Li, Xingbo and Luo, Ting and Ren, Wenqi},
  journal={IEEE Transactions on Image Processing}, 
  title={Toward Better Than Pseudo-Reference in Underwater Image Enhancement}, 
  year={2025},
  volume={34},
  number={},
  pages={6168-6179},
  month={Sep.},
  doi={10.1109/TIP.2025.3611138}
  }

@INPROCEEDINGS{zhang2024atlantis,
  author={Zhang, Fan and You, Shaodi and Li, Yu and Fu, Ying},
  booktitle={2024 IEEE/CVF Conference on Computer Vision and Pattern Recognition (CVPR)}, 
  title={Atlantis: Enabling Underwater Depth Estimation with Stable Diffusion}, 
  year={2024},
  volume={},
  number={},
  pages={11852-11861},
  doi={10.1109/CVPR52733.2024.01126}
}

@ARTICLE{jiang2025degradationiqa,
  author={Jiang, Qiuping and Gu, Yuese and Wu, Zongwei and Li, Chongyi and Xiong, Huan and Shao, Feng and Wang, Zhihua},
  journal={IEEE Transactions on Image Processing}, 
  title={Deep Underwater Image Quality Assessment With Explicit Degradation Awareness Embedding}, 
  year={2025},
  volume={34},
  number={},
  pages={1297-1310},
  month={Feb.},
  doi={10.1109/TIP.2025.3539477}
}

@ARTICLE{guo2022subjectiveiqa,
  author={Guo, Pengfei and He, Lang and Liu, Shuangyin and Zeng, Delu and Liu, Hantao},
  journal={IEEE Transactions on Multimedia}, 
  title={Underwater Image Quality Assessment: Subjective and Objective Methods}, 
  year={2022},
  volume={24},
  number={},
  pages={1980-1989},
  month={Apr.},
  doi={10.1109/TMM.2021.3074825}
}

@inproceedings{zhu2017unpaired,
  title={Unpaired image-to-image translation using cycle-consistent adversarial networks},
  author={Zhu, Jun-Yan and Park, Taesung and Isola, Phillip and Efros, Alexei A},
  booktitle={Proceedings of the IEEE international conference on computer vision},
  pages={2223--2232},
  year={2017},
  month={Oct.}
}

@article{raveendran2021underwater,
  title={Underwater image enhancement: a comprehensive review, recent trends, challenges and applications},
  author={Raveendran, Smitha and Patil, Mukesh D and Birajdar, Gajanan K},
  journal={Artificial Intelligence Review},
  volume={54},
  number={7},
  pages={5413--5467},
  year={2021},
  month={Oct.},
  publisher={Springer}
}

@ARTICLE{cong2024comprehensive,
  author={Cong, Xiaofeng and Zhao, Yu and Gui, Jie and Hou, Junming and Tao, Dacheng},
  journal={IEEE Transactions on Emerging Topics in Computational Intelligence}, 
  title={A Comprehensive Survey on Underwater Image Enhancement Based on Deep Learning}, 
  year={2026},
  volume={10},
  number={4},
  pages={2783-2801},
  month={Aug.},
  doi={10.1109/TETCI.2026.3668671}
}

@article{mohsan2023recent,
  title={Recent advances, future trends, applications and challenges of internet of underwater things (iout): A comprehensive review},
  author={Mohsan, Syed Agha Hassnain and Li, Yanlong and Sadiq, Muhammad and Liang, Junwei and Khan, Muhammad Asghar},
  journal={Journal of Marine Science and Engineering},
  volume={11},
  number={1},
  pages={124},
  year={2023},
  month={Jan.},
  publisher={MDPI}
}

@ARTICLE{saoud2024seeing,
  author={Saad Saoud, Lyes and Elmezain, Mahmoud and Sultan, Atif and Heshmat, Mohamed and Seneviratne, Lakmal and Hussain, Irfan},
  journal={IEEE Access}, 
  title={Seeing Through the Haze: A Comprehensive Review of Underwater Image Enhancement Techniques}, 
  year={2024},
  volume={12},
  number={},
  pages={145206-145233},
  month={Sep.},
  doi={10.1109/ACCESS.2024.3465550}
}

@INPROCEEDINGS{fabbri2018enhancing,
  author={Fabbri, Cameron and Islam, Md Jahidul and Sattar, Junaed},
  booktitle={2018 IEEE International Conference on Robotics and Automation (ICRA)}, 
  title={Enhancing Underwater Imagery Using Generative Adversarial Networks}, 
  year={2018},
  volume={},
  number={},
  pages={7159-7165},
  doi={10.1109/ICRA.2018.8460552}
}

@ARTICLE{jiang2024towards,
  author={Jiang, Qiuping and Yi, Xiao and Ouyang, Li and Zhou, Jingchun and Wang, Zhihua},
  journal={IEEE Transactions on Circuits and Systems for Video Technology}, 
  title={Toward Dimension-Enriched Underwater Image Quality Assessment}, 
  year={2025},
  volume={35},
  number={2},
  pages={1385-1398},
  month={Feb.},
  doi={10.1109/TCSVT.2024.3466925}
}

@ARTICLE{zheng2022uif,
  author={Zheng, Yannan and Chen, Weiling and Lin, Rongfu and Zhao, Tiesong and Le Callet, Patrick},
  journal={IEEE Transactions on Image Processing}, 
  title={{UIF}: An Objective Quality Assessment for Underwater Image Enhancement}, 
  year={2022},
  volume={31},
  number={},
  pages={5456-5468},
  month={Aug.},
  doi={10.1109/TIP.2022.3196815}
}

@article{jiang2022underwater,
  title={Underwater image enhancement quality evaluation: Benchmark dataset and objective metric},
  author={Jiang, Qiuping and Gu, Yuese and Li, Chongyi and Cong, Runmin and Shao, Feng},
  journal={IEEE Transactions on Circuits and Systems for Video Technology},
  volume={32},
  number={9},
  pages={5959--5974},
  year={2022},
  month={Apr.},
  publisher={IEEE}
}
